# Advancing Molecular Machine Learning Representations with Stereoelectronics-Infused Molecular Graphs

Daniil A. Boiko,[1] Thiago Reschützegger,[2] Benjamin Sanchez-Lengeling,[3,4,5] Samuel M. Blau,*[6] Gabe Gomes*[1,7,8,9]

1. Department of Chemical Engineering, Carnegie Mellon University, Pittsburgh, PA 15213, USA
2. Department of Chemical Engineering, Federal University of Santa Maria, Santa Maria, RS, Brazil
3. Google DeepMind, Cambridge, MA, USA (*previous affiliation, where most of this work was done*)
4. Department of Chemical Engineering and Applied Chemistry, University of Toronto, Toronto, ON M5S 3E5, Canada (*current affiliation*)
5. Vector Institute for Artificial Intelligence, Toronto, ON, Canada (*current affiliation*)
6. Energy Technologies Area, Lawrence Berkeley National Laboratory, Berkeley, CA 94720, USA
7. Department of Chemistry, Carnegie Mellon University, Pittsburgh, PA 15213, USA
8. Machine Learning Department, Carnegie Mellon University, Pittsburgh, PA 15213, USA
9. Wilton E. Scott Institute for Energy Innovation, Carnegie Mellon University, Pittsburgh, PA 15213, USA

** corresponding authors: smblau@lbl.gov, gabegomes@cmu.edu*

## Abstract

Molecular representation is a critical element in our understanding of the physical world and the foundation for modern molecular machine learning. Previous molecular machine learning models have employed strings, fingerprints, global features, and simple molecular graphs that are inherently information-sparse representations. However, as the complexity of prediction tasks increases, the molecular representation needs to encode higher fidelity information. This work introduces a novel approach to infusing quantum-chemical-rich information into molecular graphs via stereoelectronic effects, enhancing expressivity and interpretability. Learning to predict the stereoelectronics-infused representation with a tailored double graph neural network workflow enables its application to any downstream molecular machine learning task without expensive quantum chemical calculations. We show that the explicit addition of stereoelectronic information significantly improves the performance of message-passing 2D machine learning models for molecular property prediction. We show that the learned representations trained on small molecules can accurately extrapolate to much larger molecular structures, yielding chemical insight into orbital interactions for previously intractable systems, such as entire proteins, opening new avenues of molecular design. Finally, we have developed a web application (simg.cheme.cmu.edu) where users can rapidly explore stereoelectronic information for their own molecular systems.

# Introduction

Molecular representation is a cornerstone in chemistry.[1,2] Following chemists' intuition, skeletal structures became the chemical *lingua franca*. They allow us to capture the wide diversity of (mostly organic) molecules, while preserving simplicity and making it easier for humans to recognize common patterns. In addition to influencing the way chemistry is thought about and described,[3] these representations are powering advances in molecular machine learning (ML), which has been used for various applications.

One of the most successful applications of ML in such settings is the prediction of molecular properties, which is at the core of chemical, biological, and material sciences. From the discovery of materials for solar panels[4] to the record-setting development of a new drug,[5,6] molecular ML has significantly impacted modern science by enabling fast inference. The performance of the ML models is strongly connected to the underlying molecular representation, arguably the most critical factor for their success.[7] Current standard molecular representations encompass various approaches: global descriptors,[8] strings that translate the structure into a token sequence,[9] graphs that encode covalent bonding information,[10,11] providing topological information (**Figure 1a**). There are also approaches to infuse spatial features into latter representations, providing structural information.[12,13] Recent approaches have also shown how combining force field representations with electronic structure descriptions can enable simultaneous prediction of both structural and electronic properties.[14] Some important representations[15] extend or improve on Coulomb matrix eigenvalues molecule representations.[16,17] Finally, selection of a representation is closely tied to corresponding model architecture and data processing. Many approaches were developed to enforce certain symmetry considerations; for example, by averaging across multiple local coordinate systems[18] (a similar approach was used in AlphaFold 3[19]), or making changes at the model level.[13]

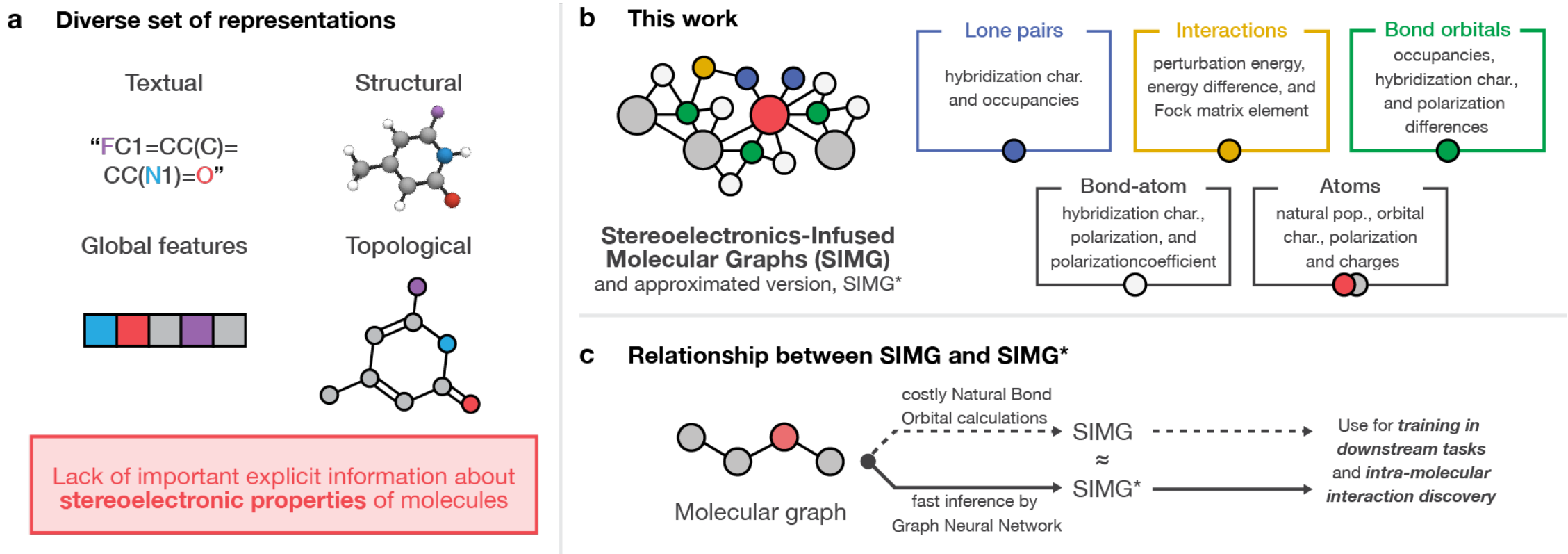

**Figure 1. Common molecular representations and overview of our approach**. **a**, Various popular molecular representations for machine learning strategies. **b,** ***SIMG*** construction approach. **c,** the relationship between ***SIMG*** and ***SIMG****.

Despite the success of machine learning for molecular property predictions, the representations used in these methods are incomplete. Current widely used graph representations lack quantum-chemical priors from the electronic structure of molecules (**Figure 1a**) or have very limited interpretability. However, the computational chemistry community has developed powerful techniques for describing the quantum-mechanical nature of molecular structures by quantifying the presence and strength of orbital interactions. Such "stereoelectronic" information provides valuable fundamental chemical insight. For example, interactions between the corresponding bonding/nonbonding and antibonding orbitals are essential to our understanding of many chemical phenomena,[20] such as in protein-substrate interactions[21] and organocatalysis.[22] The value of stereoelectronic information suggests that its successful incorporation into molecular representations could improve ML model performance, as long as costly calculations of stereoelectronic features of input structures can be avoided during downstream training and inference.

This work introduces a new representation based on molecular graphs enhanced with nodes corresponding to bond orbitals, lone pairs, and interactions between them (which essentially encode relational 3D information), called ***SIMG**s* (stereoelectronics-infused molecular graphs). We describe how the ***SIMG*** representation (**Figure 1b**) can be constructed from Natural Bond Orbital[23] (NBO) analysis data and approximated with graph neural networks (***SIMG****) to enable fast predictions (**Figure 1c**). We study the benefits of using these representations as input for message-passing 2D machine learning algorithms to perform molecular property prediction and the capacity for ***SIMG****s* to identify stereoelectronic orbital interactions in systems where direct quantum chemical NBO calculations are intractable, unlocking previously inaccessible chemical insight.

# Results

In this work, our goal is to build a new molecular machine learning representation that achieves three equally important goals: I. enhances the performance of downstream models, when compared to other molecular graph representations; II. is based on our collective understanding that molecules are three-dimensional quantum mechanical objects; III. is easily human interpretable.

To develop this representation, we start with constructing these new graphs from NBO calculation results. Then, we show how this representation can be approximated with graph neural networks. The developed representation is then applied to two tasks. The first is the prediction of QM7 and QM9 targets, serving as an example of downstream applications where we compare our results with Coulomb matrix,[17] SPAHM,[15] MAOC,[24] SOAP,[25] gradient boosted decision tree model[26] on top of MegaMolBART[27] embeddings, molecular graph, and ChemProp[28] baselines. The second involves discovering graph-distant proximal orbital interactions in proteins, through which we demonstrate that our approach effectively extrapolates beyond the small molecule training data by exploiting the spatially local character of orbital interactions.

## 1. Stereoelectronics-Infused Molecular Graphs (*SIMG*s)

The proposed representation is based on Natural Bond Orbital analysis and a new heterogeneous graph topology. NBO analysis starts with providing a 3D structure of a given molecule and a valid description of its wave function.[23] NBO yields a collection of localized natural atomic orbitals, hybrid orbitals, bonding/nonbonding, and antibonding orbitals.[29] It does so by doing a series of transformations with the Fock-matrix[30] term from Hartree-Fock-derived quantum-chemistry methods,[30] including hybrid density functional approximations.[31,32] Moreover, NBO analysis quantifies interactions between filled orbitals (donors) and empty orbitals (acceptors) via second-order perturbation interactions. This strategy yields a quantitative description of the quantum-mechanical nature of electronic density delocalization, rendering itself a perfect fit for heterogeneous molecular graph representations.

***SIMG*s** are constructed using the molecule's NBO analysis data (**Figure 1b**). In contrast with standard molecular graphs, which simply represent molecules as a collection of nodes for each atom and edges for each covalent bond, we propose the inclusion of new types of stereoelectronic nodes and edges. Besides the traditional atomic node, we include lone pair nodes ($n$) and bond nodes ($\sigma$, $\sigma^*$, $\pi$, $\pi^*$), which represent electron density orbitals, and nodes which represent interactions between orbitals. Although there are works on infusing this information implicitly,[33] we chose an explicit approach to emphasize interpretability and better control the graph structure.

Furthermore, we extend the topology changes from molecular graphs to ***SIMG*s** with numerical NBO analysis information (**Figure 1b**). In particular, Natural Population Analysis provides localized electron information for each atom, which is stored in ***SIMG*s** as atomic features. Atom targets include their charge, the number of core electrons, valence electrons, and total electrons. For bond orbitals, we include atom-wise hybridization characters, bond polarization and its coefficients, and the respective values for antibonding orbitals. Orbital hybridization of nonbonding orbitals (*e.g.*, lone pairs) information is also used. Donor-acceptor interactions are represented as connections between the respective interacting nodes with corresponding numerical description. For more details see description in "Methods" section. Overall, these representations show superior results over standard molecular graphs (see Section 4 for more details) while also enhancing interpretability.

## 2. *SIMG*s surrogate with Graph Neural Networks.

Computation of required NBO data for ***SIMG*** construction can take substantial time, severely limiting the use of such graphs over large datasets of molecular structures and conformer ensembles. Moreover, for large molecules such as proteins, these calculations are simply not feasible due to the limitations in the number of basis functions that NBO can process. The main workaround is via truncation of the macromolecular structure, a strategy that severely limits how we study such systems. Therefore, we aimed to develop an approach for fast prediction of ***SIMG*** graphs solely from molecular three-dimensional structure (*e.g.*, Cartesian coordinates). We denote these approximated representations as ***SIMG****.

***SIMG**** prediction is accomplished by a neural network architecture that will be described shortly. For model training, we run DFT + NBO on the full QM9 dataset and a portion of the GEOM dataset, where further details of the DFT are provided in the Methods section. In the following sections, we discuss results with models either trained on QM9 NBO data (property prediction and architecture optimization) or GEOM NBO data (***evolver*** module, active learning, protein applications, and property prediction). We also develop our own benchmark dataset for large molecules.

### 2.1. General approach for *SIMG** construction

Interactions involving nonbonding orbitals are critical to molecular structure.[34] However, such information is absent from traditional molecular graphs, as it must be obtained from expensive quantum chemical calculations. We overcome this limitation by formulating the problem as a sequential graph construction (**Figure 2a)**.

First, a naive molecular graph is constructed from the three-dimensional input structure which is used as the input for a model which predicts the number of lone pairs and their types (see full description in the "Methods" section). Predicted lone pair information facilitates the construction of an intermediate graph, which we call an "extended molecular graph," which contains both lone pair nodes and nodes that represent $\sigma$- and $\pi$-bonds, a superior description to simply representing any covalent bond as an edge. These graphs are subsequently used as input for a second model, designed to predict electronic properties of orbital overlaps in a multitask fashion, *i.e.*, the ***SIMG**** graphs.

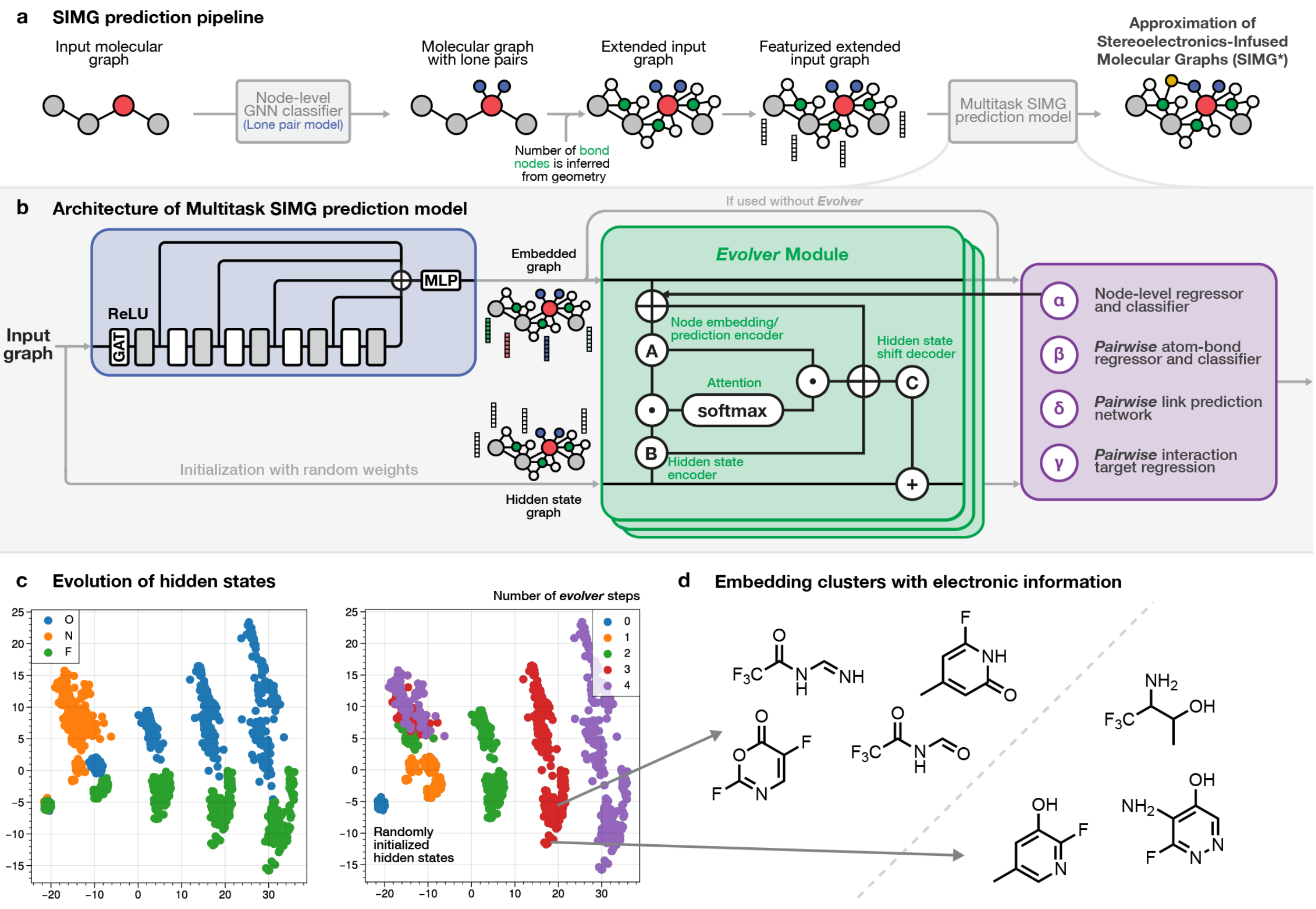

**Figure 2. Approach to *SIMG** construction. a,** ***SIMG****s are constructed from molecular graphs using two models (in the example molecule, hydrogens are not shown). **b,** Architecture of the multitask model, including the ***evolver*** module. **c,** Evolution of the hidden states marked by atom type and step number. **d,** Two clusters of fluorine lone pairs distinguishing the presence or absence of C=O bonds.

### 2.2. Architecture of the multitask model.

The multitask model takes an extended molecular graph as an input and passes it through a GNN-based encoder (**Figure 2b**).[11] The encoder uses graph attention layers and includes skip-connection to address oversmoothing issues (see details in the Methods section). The output of the encoder is a set of embeddings for each of the nodes in the graph.

We employ a two-pronged approach to get ***SIMG****s from these embeddings because of the following challenge: we explicitly introduce lone pair and bond nodes that may have different targets. For example, furane has an oxygen atom with two lone pairs: a pure *p*-orbital and a $sp^{\lambda}$-hybrid lone-pair. The latter can participate in the formation of hydrogen bonds and other non-covalent interactions, but, in the input extended molecular graph their features are identical. We introduce two approaches to solve this problem:

1. **Assign features that make it possible to differentiate initially feature-identical nodes.** This is accomplished by making a lone pair model to predict mostly *p*- or mostly *s*-character of the lone pair (see the "Methods" section for more details).
2. **Introduce random hidden states and perform multiple updates of these hidden states.** This is carried out using five blocks of the ***evolver*** module (**Figure 2b**) that uses node embeddings in intermediate predictions of node targets to update these hidden states in combination with a permutation-invariant loss function (see the "Methods" section for a detailed description and Figure S2 in the Supporting Information).

Both approaches work better than the baseline model without any differentiating information (see Supporting Information, Figures S4-5 for comparison). Although the ***evolver*** approach is a slightly less efficient than assigning differentiating features, it is much more general.

The operations performed by the ***evolver*** module are based on iterative updates of the hidden states. Therefore, we can study how embeddings change with iterations. **Figure 2c** shows a PCA map for the node hidden states, colored by element and time step, respectively (see Figure S3 for more specific examples). Initially essentially indistinguishable, the model slowly changes embeddings to spread them apart. Moreover, elements like fluorine have two different clusters. Inspection of these clusters revealed an interesting outcome (**Figure 2d**): for fluorine, these clusters differ solely based on the presence of a C=O double bond in the molecule. Finally, in comparison to a vanilla GNN (see Supporting Information, Figure S9), we see a much better separation with our architecture.

### 2.3. Active learning of *SIMG*s based on epistemic uncertainty estimation

A primary goal of our work is to enable predictions for a set of elements covering common biomolecules. One of the datasets with enough chemical diversity in terms of elements, molecular size, and conformations is GEOM.[35] However, its immense scale makes running DFT + NBO calculations unreasonably costly. Instead, we decided to perform training in an active learning manner (**Figure 3a**).

The proposed methodology heavily relied on the estimation of uncertainties in the prediction of ***SIMG***s. These uncertainties can be categorized into two types: aleatoric uncertainty and epistemic uncertainty, which has great importance in this context.[36,37] While there exist theoretically rigorous approaches such as Bayesian neural networks for uncertainty estimation,[38] these methods are notorious for their challenging training requirements. Thus, in practical implementation, we adopted an ensemble of models and computed the variance of their predictions, that has proven to yield good results.[39]

In this work, we developed the following approach. First, a uniform sample of 29,500 structures from GEOM was chosen and NBO calculations were performed. We used this data to train a lone pair prediction model and an ensemble of three multitask ***SIMG**** prediction models. Next, we performed lone pair model inference and then featurized extended molecular graph construction for each structure in the GEOM dataset. We divided the dataset into 295 parts and selected 10,000 molecules from each part to pass through the models in the ensemble. For each molecule and target, we calculated the variances of the ensemble models predictions. We then selected the 1,000 molecules with the highest variances for each target, removed duplicates, and ran DFT + NBO calculations on the new sample. As the procedure was done for every target, we ended up with 20,000–30,000 molecular structures to run DFT + NBO on each time.

The dataset additions throughout the iterations of the active learning procedure unveiled multiple observations, as shown in **Figure 3b** (see Figure S6 for more details). During the first iteration, the ensemble showed a preference for selecting a substantial number of molecules containing boron, initially underrepresented in the dataset. Additionally, molecules with bromine and iodine, as well as molecules containing both iodine and sulfur, were chosen. In subsequent iterations, there was an increased focus on selecting molecules containing mercury and arsenic. Notably, in the final step of the active learning procedure, the ensemble collected numerous molecules with a B-Cl pair. Furthermore, we conducted an analysis of the relative positions of rings within the molecule (refer to **Figure 3b**). During the final iteration, our ensemble-based approach demonstrated a capability to selectively target compounds featuring adamantane structures. This ability to discern and focus on specific subsets of compounds showcases the approach's efficacy in effectively navigating the chemical space.

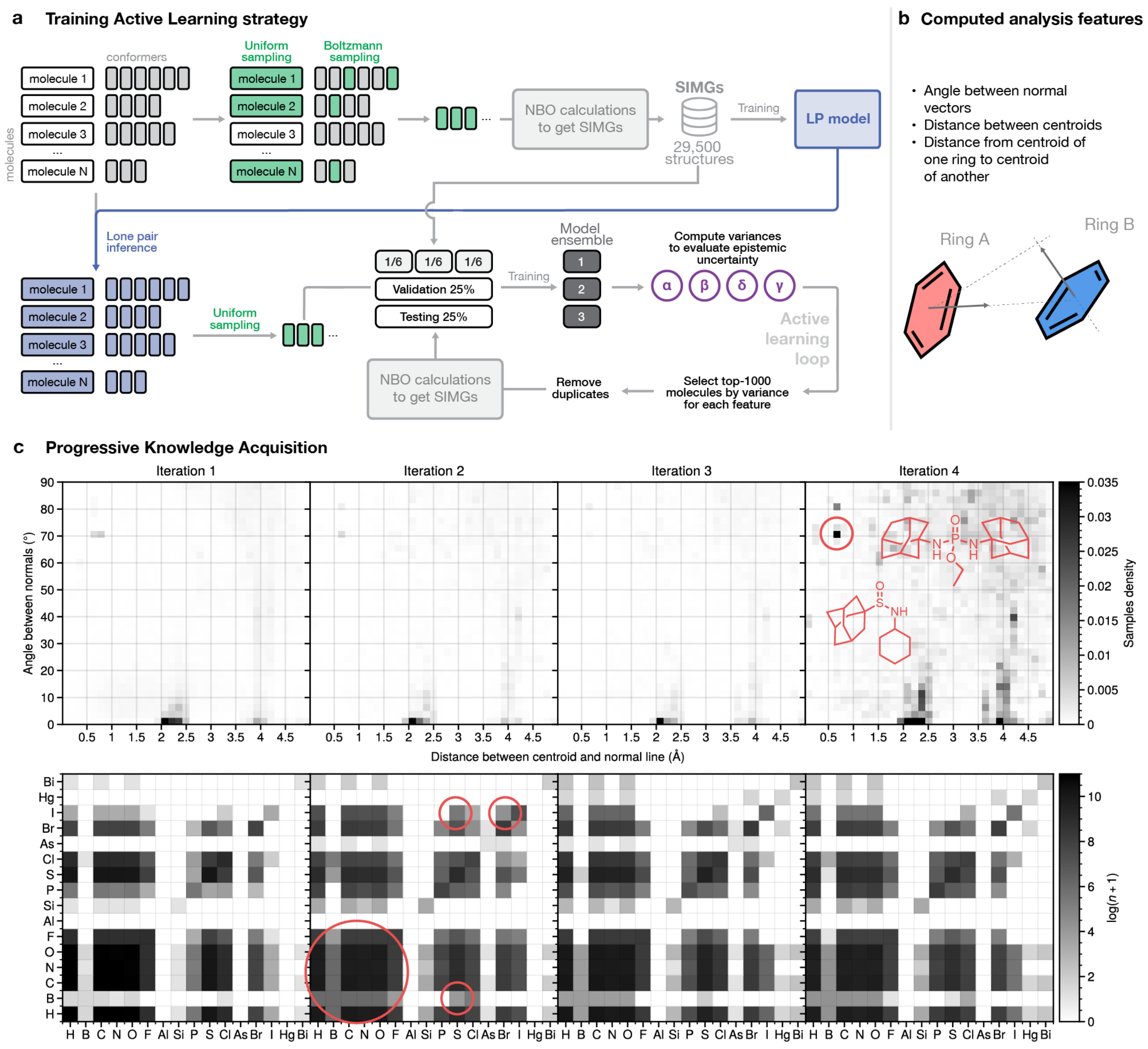

**Figure 3. Active learning approach to select additional NBO data to generate for the model. a,** Overview of the approach. **b,** Features, calculated to analyze chemical space shifts and changes of the data acquisition through iterations of the active learning procedure.

## 2.4. Prediction performance

**Figure 4** highlights the successful approximation of ***SIMG****s utilizing graph neural networks. The model demonstrated outstanding performance in classifying the quantity and types of lone pairs at the node level, as evidenced by the confusion matrix presented in **Figure 4a**, while being able to reconstruct the ground-truth extended graph in 98% of the cases.

Node-level tasks included the prediction of properties for atoms, lone pairs, and bonds. Remarkable performance was observed in atom-related tasks with excellent $R^2$ scores achieved. For lone pair-related tasks, particularly those involving *s*- and *p*-character, excellent predictive scores were attained. Although slightly lower scores were observed for the *d*- prediction task due to the limited number of *d*-block data points, predicted values remained sufficiently reliable. Successful predictions were also observed for lone pair occupancies. For bond-related tasks, occupancies were predicted with favorable scores. Additionally, predictions related to hybridization characters and polarizations exhibited notable accuracy with the exception for *d*-character, simply due to the smaller number of *d*-orbital containing elements in the dataset.

Some prediction targets for bonds are different for each atom, such as hybridization characters and polarizations, and polarization coefficients. Exceptional performance was achieved for all targets except *d*-characters for the previously discussed reasons.

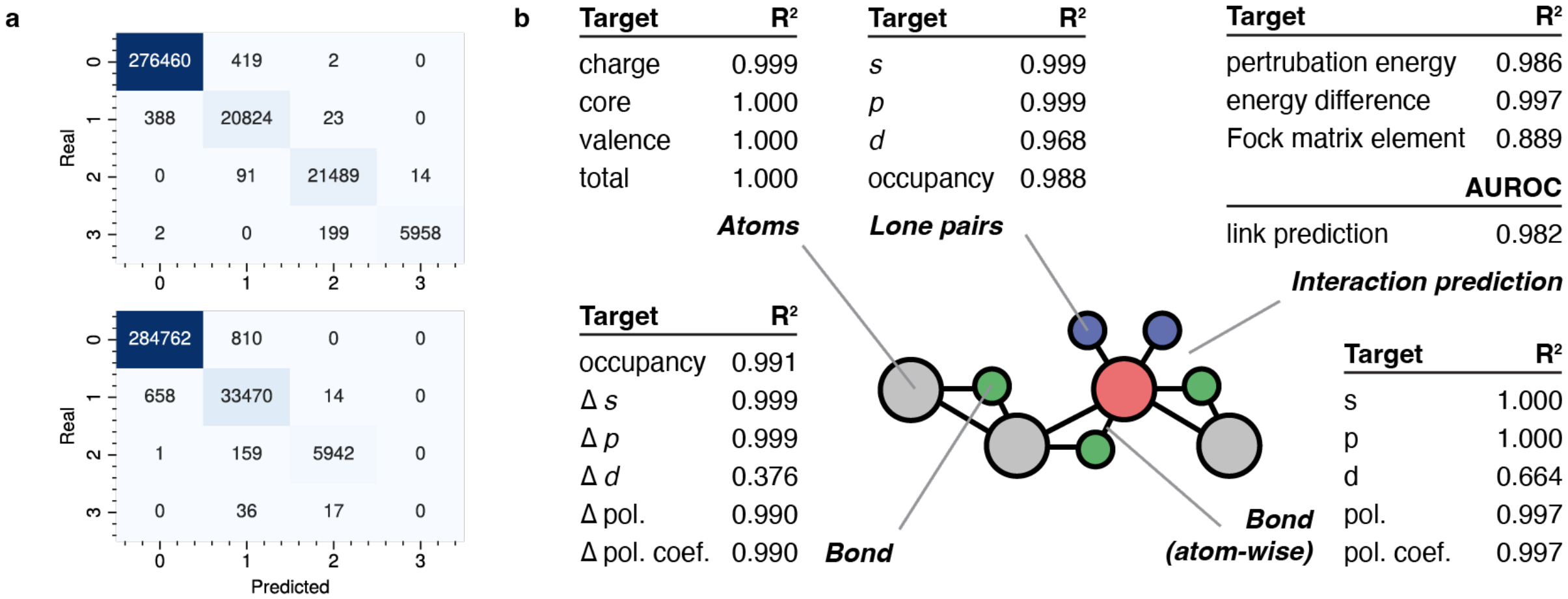

**Figure 4. Quality of *SIMG** predictions. a,** Confusion matrix for lone pair count, and type prediction, respectively. **b,** Prediction performance for the second model.

As prediction of second-order orbital interactions is a classification problem, we used AUROC (area under the receiver operating characteristic curve) and mean AUROC. The value of AUROC is equal to 0.982, which is a good result for such complex tasks. The value for the mean AUROC of 0.979 is only slightly smaller. Finally, the properties of these interactions were also predicted; the only decline in performance is observed for the values of Fock matrix elements (**Figure 4b**).

### 3. ***SIMG****s enable rapid prediction of orbital interactions for previously inaccessible molecules.

Stereoelectronic interactions, as derived from NBO data and subsequently represented by and predicted with ***SIMG*****s**, assume a crucial role in the stability of large molecules.[40–42] Notably, while spatial structures of proteins primarily rely on the presence of hydrogen bonds for stability, an array of weaker orbital interactions collectively contributes significantly to the overall stabilization energy. Unfortunately, even the most recent version of the NBO program struggles to perform analysis atop DFT simulations with more than ~3000 basis functions, making them infeasible for most macromolecules. However, owing to the spatially localized nature of stereoelectronic effects, we hypothesized that an ***SIMG**** model trained on NBO data of small molecules in the GEOM dataset could accurately predict orbital interactions in much larger structures, unlocking previously inaccessible chemical insight.

To assess the model's effectiveness in identifying particular orbital interactions within large molecules, particularly proteins, we focused on a subset of the Potassium Channel KcsA-Fab complex in a low concentration of $K^+$ (PDB code 1k4d).[43] The specific features of interest were the orbital interactions in an $\alpha$-helix involving the *p*-rich lone pair electrons of the amide oxygen atom ($n_O$) and the antibonding $\pi$-orbital of the adjacent carbonyl group ($\pi^*_{CO}$), as shown in **Figure 5d**. Such $n_O \rightarrow \pi^*_{CO}$ interactions were previously reported in this protein and were predicted through NBO analysis, serving as a basis for our evaluation.[44]

In the context of the entire protein structure, we built Ramachandran plots as presented in **Figure 5a**. Specifically, we focused on identifying the dihedral angles associated with selected subgraphs where the potential presence of $n_O \rightarrow \pi^*_{CO}$ interactions could occur. These dihedral angles are defined by $\phi(C'_{i-1}–N_i–C^{\alpha}_i–C'_i)$ and $\Psi(N_i–C^{\alpha}_i–C'_i–N_{i+1})$. Our model successfully predicted the aforementioned orbital interactions in most instances, as depicted in **Figure 5a**. Finally, model predictions of second-order orbital interaction are closely related to its ground truth (**Figure 5c**).

Furthermore, we aimed to assess the model's performance in capturing and comprehending graph-distant proximal orbital interactions, which are far apart when traversing the bonding graph but close together in 3D space. To achieve this, we quantified the ***SIMG**** predicted interactions and performed a comparative analysis with ground-truth NBO calculations, using F1-score as shown in **Figure 5b**. Observations highlighted a positive correlation between the model's performance and the abundance of interaction samples. We note that when the distance between atoms was below 2.8 Å and the bond graph distance was under 4, there was a substantial increase in the number of existing interactions. This trend contributed to the model's enhanced performance in predicting interactions within these ranges. Critically, we note that the available training data with instances of more spatially distant interactions (atom distances >2.8 Å, graph distances >4) is scarce.

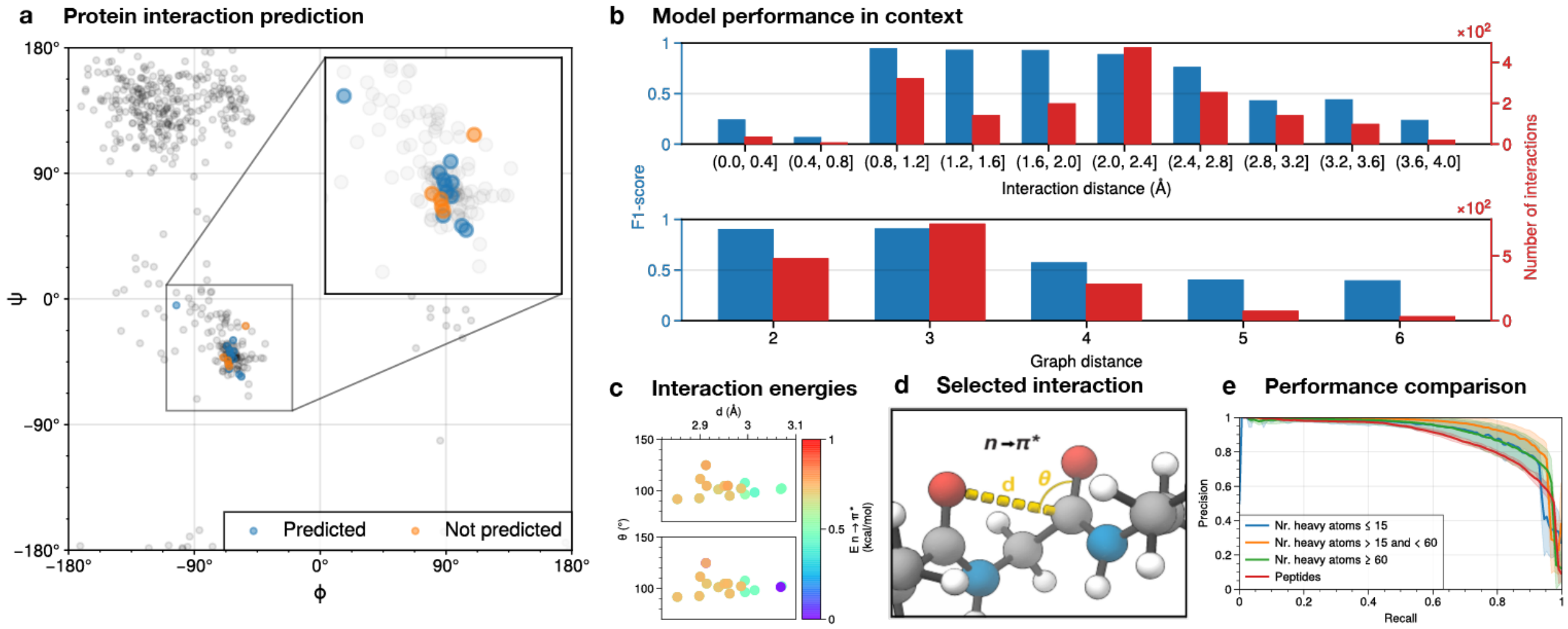

**Figure 5. Assessment of Model Performance in Identifying Structural Features and Graph-distant Proximal Orbital Interactions in Proteins. a,** Ramachandran plot illustrating the identification of structural features, specifically the selected subgraphs and associated dihedral angles ($\phi$ and $\Psi$) related to potential $n_O \rightarrow \pi^*_{CO}$ interactions in the $\alpha$-helix of the Potassium Channel KcsA-Fab complex. **b,** Model performance evaluation for capturing graph-distant proximal orbital interactions using the F1-score metric. (Graph distance histogram was truncated for better readability). **c,** Strong correlation is observed between predicted and actual energy values associated with the identified interactions. **d,** Visualization of a $n_O \rightarrow \pi^*_{CO}$ contact within the $\alpha$-helix structure, supported by previous reports and NBO analyses. **e,** Precision-Recall curves comparing the performance of the model for molecules of varying sizes, from small molecules to peptides.

To see how well our models perform for large structures in general, we created an additional set of large molecules (neutral, closed-shell, truncated peptides approaching the limit of computational feasibility with the quantum chemistry methodologies employed in this work) and performed NBO calculations on them to get ***SIMG***s. These ***SIMG***s were then compared with predicted ***SIMG*s***. We emphasize that while DFT + NBO calculations on these structures required hours or even days to run, ***SIMG**** prediction on identical structures takes seconds. **Figure 5e** shows that the model successfully captures the stereoelectronic interactions without significant differences in performance when compared with smaller molecules. Moreover, the only significant difference between a large molecule and an example from the GEOM dataset is bond graph distance, not observable spatial distance. We compared the model's performance for cases with various bond graph distances calculated with Dijkstra's algorithm and have not observed any correlation between them. Aggregated metrics also reveal no significant difference, validating the extrapolatory capacity of our ***SIMG**** model predictions.

## 4. *SIMG* and *SIMG** representations improve prediction performance of downstream message-passing 2D molecular machine learning models

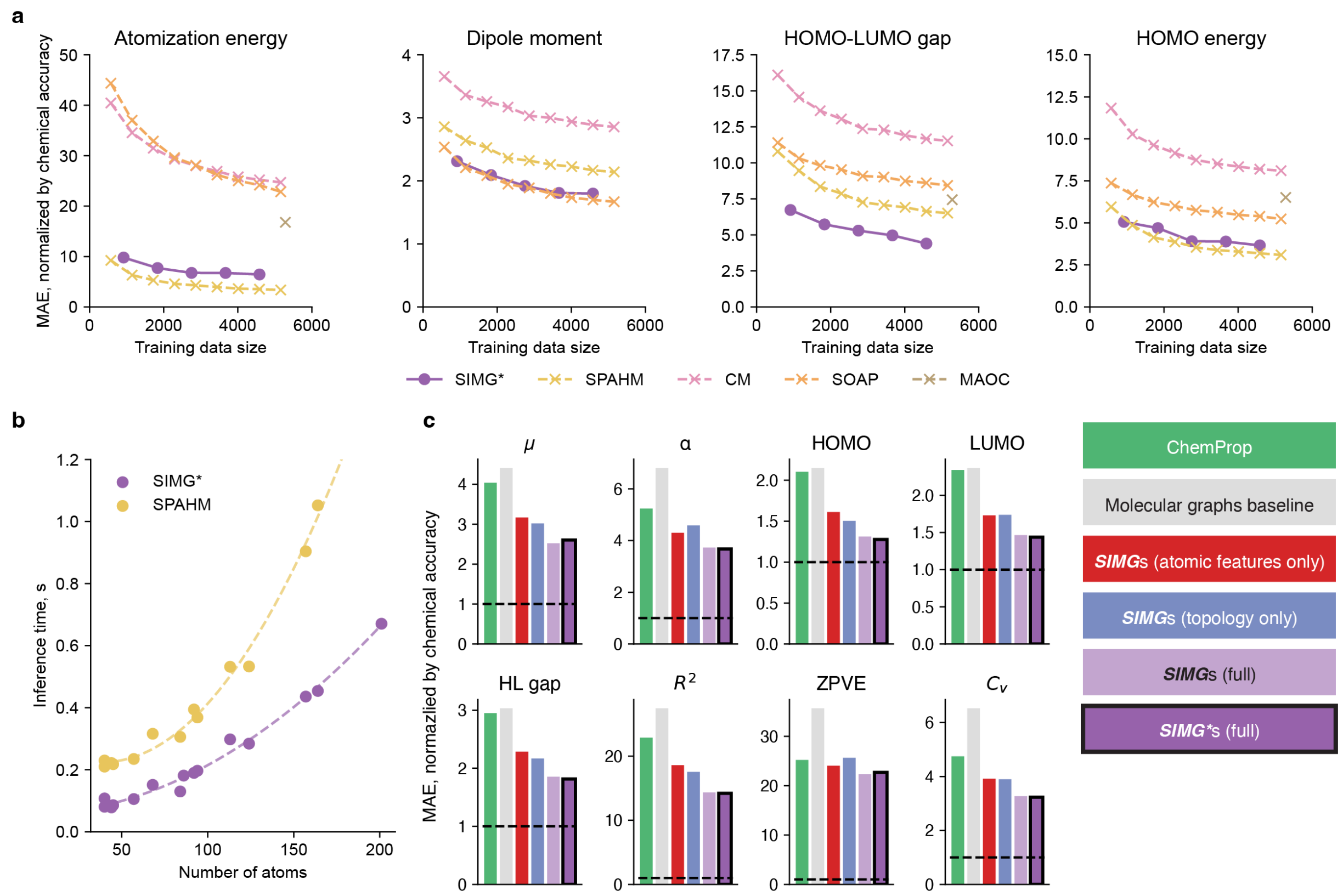

**Figure 6.** Property prediction performance of models employing different molecular representations. **a,** Property prediction performance comparison as a function of training data size with various representations (including physics-informed ones) when training a kernel ridge regression or a message-passing 2D GNN (for ***SIMG****) on the QM7 dataset, where lower mean absolute error is better. Metrics are normalized by chemical accuracy. **b,** Comparison of representation generation speed between SPAHM and ***SIMG****. **c,** Property prediction performance comparison when training on the QM9 dataset with naïve molecular graphs, molecular graphs using ChemProp, and ***SIMG***s, where we distinguish between ***SIMG***s with only atomic features vs. only topology vs full ***SIMG***s vs full ***SIMG****s. Metrics (mean absolute error) are normalized by chemical accuracy. Each subplot represents one of the targets from QM9 dataset, bar chart elements correspond to our models and baselines. The horizontal dashed line indicates chemical accuracy.

Finally, we evaluate our representation's efficacy in downstream tasks. First, we compare ***SIMG**** representation for training a message-passing 2D GNN property prediction model with different physics- and geometry-informed representations for training a kernel ridge regression (as used in corresponding publications) on the QM7 dataset (**Figure 6a**). Our representation outperforms all others for HOMO-LUMO gap prediction, matches the performance of the best representation (SOAP) for dipole moment prediction, is barely worse than the best representation (SPAHM) on HOMO energy prediction and atomization energy prediction. We also note that some of these methods performed extensive hyperparameter tuning for the QM7 prediction task, while no such tuning was performed for ***SIMG****.

Second, we compared the speed of representation generation with SPAHM, given that it is another physics-informed method which showed strong performance on QM7. As can be seen from **Figure 6b**, ***SIMG**** generation is consistently faster than SPAHM generation across a large set of molecule sizes, and in particular the performance gap grows when scaling to quite large molecular systems.

Finally, we perform additional comparative analysis when training on a larger-scale dataset, the widely used QM9. We assessed the influence of graph augmentation by comparing three distinct graph structures (**Figure 6c**): *i)* molecular graphs with additional features as done in ChemProp (default hyperparameters);[28] *ii*) standard molecular graphs; *iii*) ***SIMG***s having original molecular graph topology but extended atom feature set; *iv*) ***SIMG***s incorporating topological structure without adding new features; and *v*) complete ***SIMG***s. Through this comparative analysis, we were able to distinguish effects caused by the incorporation of additional features and changing graph topology. Furthermore, the evaluation was conducted utilizing both ground-truth ***SIMG***s and their learned ***SIMG**** counterparts generated by our model (*vi*).

**Figure 6c** shows the performance of the different graph representations using the same architecture type — message-passing 2D GNN model. ***SIMG***s with either atomic features or topology provide a comparable improvement over both the molecular graph and ChemProp baselines (with the exception of ZPVE for the latter). Full ***SIMG***s yield an even bigger improvement, especially for tasks such as dipole moment prediction. Finally, the learned representation ***SIMG**** displays virtually the same performance as the original ***SIMG*** while avoiding costly DFT + NBO calculations during downstream model training. Moreover, in terms of absolute values for many targets such as $\mu$ and HL gap, our learned representation brings corresponding models much closer to the chemical accuracy target. We further compared against a gradient-boosted decision tree model on top of MegaMolBART embeddings, but the performance was substantially worse than the other baselines shown here (see Supporting Information).

The provided results confirm a significant improvement in the performance of message-passing 2D GNN property prediction models using our developed representation. However, it must be noted that such 2D GNN models perform substantially worse for the tasks shown here than 3D GNNs using equivariant architectures. While we believe that ***SIMG****s can readily be employed as an input to improve 3D GNN models for property prediction and interatomic potentials, and for other tasks such as molecular conformer generation and protein folding, these topics are beyond the scope of the present work.

# Discussion

This work proposed stereoelectronics-infused molecular graphs (***SIMG***s): a new type of molecular machine learning representation that incorporates quantum-chemical interactions beyond classic Lewis structures, enhancing interpretability. ***SIMG***s substantially improve performance in downstream ML applications, such as message-passing 2D GNNs for property prediction. Moreover, we described how our representation can be approximated with graph neural networks, yielding ***SIMG****s. We further demonstrated that the ***SIMG**** prediction model can be trained on small molecule data and then accurately predict the representation for entire proteins, providing rapid access to chemical insight from orbital interaction information where DFT calculations are very expensive or completely intractable.

**Model limitation and recommendations**

In our representations, elements are one-hot encoded, so to add another element, one needs to collect an extended dataset of NBO analysis data for structures including the new element. This could be circumvented by using the physical properties of elements as features, but it requires further research. The present work is also constrained to overall neutral, closed-shell molecular structures, but we are in the process of extending the approach to variable charge and open-shell species.

**Broader impacts**

Molecular machine learning is a critical component of pipelines for drug and material discovery, catalyst optimization, and a valuable tool for studying complex biochemical processes. Infusion of quantum mechanical features into graph representations for molecular ML will increase trust in these algorithms, contribute to increased interpretability of the models, and open new opportunities to research the relationship between electronic structure and molecular properties. The predicted orbital information can be used to analyze chemical reactivity in a wide range of systems.[45–48]

# Methods

## 1. Data collection

Given the optimized structures for each molecule of the QM9 dataset,[49] we conducted single-point calculations at the $\omega$B97M-V[50]/def2-SVPD level of theory.

We used Q-Chem 6.0.1[51] interfaced with NBO 7.0[52] and performed the calculations via the high-throughput workflow infrastructure implemented in the Materials Project open-source software codes Pymatgen,[53] Custodian,[54] and Atomate,[55] resulting in targets for atom, bond, lone pairs, and orbital interactions, illustrated in **Figure 1b**. These are described as follows:

**Atom features.** The performed natural atomic orbital analysis returns localized electron information for each atom. Atom targets include their charge, the number of core electrons, valence electrons, and total electrons. Even though NBO analysis provides Rydberg orbitals, we did not keep them as a model's target due to the controversy in the physical meaning.

**Bond features.** In the context of localized natural bond orbitals, bonds are simply a combination of the orbitals from each atom. For that reason, the NBO analysis data provides atom-wise *s*, *p*, *d*, and *f* characters, polarization, polarization coefficient, and the respective values for antibonding orbitals. Occupancy for bonding and antibonding orbitals are the only bond-specific target from the original data. In total, there are a total of twenty-six targets.

**Lone pair features.** Orbital hybridization is described by the *s*, *p*, *d*, and *f* characters. Also, the NBO analysis provides information about its occupancy, summing up to five targets.

**Orbital (2^nd^-order) interactions.** These represent the interactions between donor and acceptor orbitals. Donors, represented by lone pairs $n$, $\sigma$, and $\pi$ bonds, are electron-rich orbitals, while acceptors, represented by $\sigma^*$, and $\pi^*$ anti-bonds are electron deficient. In practical terms, our ground truth graph represents one donor-acceptor interaction as a connection between the respective nodes. The NBO analysis quantified these interactions by the perturbation energy, energy difference, and the Fock-matrix element, with a total of three targets.

## 2. Model implementation and training

All models were implemented using PyTorch framework.[56] Model training and metric collection was implemented using PyTorch Lightning framework.[57] PyTorch Geometric was used for graph neural networks.[58] Code is available at https://github.com/gomesgroup/simg.

## 3. Lone pair prediction

**Tasks.** Although many heuristics can be used to define the number of lone pairs, we argue that a data-driven model is more well-suited in this case since it can interpolate between different contexts. With this in mind, we built a neural network capable of predicting each atom's number of lone pairs and their types. Such types were used to distinguish lone pairs of the same atom, since there are possible differences in their NBO data. Therefore, we determined the types by an analytical threshold relating *s*- and *p*-characters, expressing the conjugation likelihood of a given lone pair. The threshold is defined by the inequality below:

$$p_{character} - s_{character} > 80$$

Indeed, this relationship expresses the conjugation likelihood of a lone pair. In that sense, we train the neural network to predict how many lone pairs satisfy the threshold.

**Graph encoder.** Both tasks were tackled in tandem with a mapping function modeled as a GNN. To mitigate the over-smoothing problem, we used multiple aggregation functions through the message-passing scheme,[59] along with residual connections.[60,61] The encoder is constructed by stacking several propagation layers followed by a ReLU activation function. Node embeddings are then concatenated with a residual connection from the input graph. Finally, the result is forwarded to a multilayer perceptron (MLP), which, in this case, is composed of two linear layers separated by a ReLU activation layer.

**Training.** As a design choice, we framed both tasks as node-level classification issues, in which each class represents the number of lone pairs and how many satisfy the threshold, respectively. Therefore, we used the sum of the cross-entropy loss of each task as the loss function.

## 4. Property and interaction prediction model

### 4.1. General remarks

The neural network architecture comprises two parts: the node encoder and a group of multiple separate MLPs. The latter make predictions using the embeddings from the encoder part, but multiple preprocessing steps may be conducted depending on the task.

The encoder is constructed by stacking multiple graph neural network blocks and concatenating the outputs of each block. A block comprises a graph attention layer and a ReLU activation layer.[62] None of the dropout or batch normalization layers was used. Concatenated outputs were then passed into the MLP network with one single layer to construct the node embedding.

This encoder architecture is designed to tackle the over-smoothing issues of graph neural networks.[63] The problem arises when multiple graph neural network layers are stacked, making the computational graphs nearly identical. Over-smoothing might not be a severe issue for graph-level tasks, but it is a significant issue in performing node-level tasks. There is a wide variety of approaches to solve it.[60,61] In this work, we concatenate outputs of intermediate layers tackling both the over-smoothing and vanishing gradients difficulties.

All MLPs for separate tasks follow the same architecture: one linear layer, the ReLU activation, the batch normalization layer, and one final linear layer. The following sections describe input preparation and loss functions for these networks.

$$L = L_\alpha + L_\beta + L_\gamma + L_\delta$$

**Atom, lone pair, and bond nodes**. The most straightforward difficulty to solve is the prediction of targets for individual nodes. Here, for all types of nodes, only one network is used. The loss function was defined as a sum of separate losses for each node type. The mean squared error (MSE) loss was used for all features except orbital characters of lone pairs. Orbital character prediction was optimized with a cross-entropy loss function. MSE was also the key metric for this type of task. $R^2$ scores were also recorded.

$$L_\alpha = MSE(\alpha(x), y) + BCE(\alpha(x), y)$$

**Atom-wise bond target prediction.** Some bond features are related to each of the atoms. To keep permutation invariance, it was impossible to predict them in the previous step. The task was solved by concatenating embeddings of the atom in question and the corresponding bond, and then passing it into the MLP. Polarization value prediction was optimized with MSE loss. Orbital characters were optimized with cross-entropy loss. Similar to the previous section, MSE and the $R^2$ score were used to control the training.

$$L_\beta = MSE(\beta(x), y) + BCE(\beta(x), y)$$

**Link prediction approach.** Orbital interactions data is not available directly from the molecular structure, so it should be predicted first. Therefore, the problem was formulated as a link prediction task, essentially a classification problem. “Positive examples” (i.e., cases where there is an interaction) were taken from the original dataset, while “negative examples” were sampled from other possible combinations of bonds and lone pairs. Moreover, the direction is essential as (in our case) it describes the donor-acceptor pair but not vice versa. Input data consisted of concatenated embeddings of corresponding nodes and dynamically calculated pairwise features. We performed the training with binary cross-entropy loss. Standard classification metrics were calculated for accuracy, precision, recall, F1-score, and area under the receiver operating characteristic curve (ROC AUC).

$$L_\gamma = BCE\left(\gamma\left(x_i, x_j, p_{i,j}\right), y\right) = \sum_{i,j}^{n} -y \log(\gamma(x_i, x_j, p_{i,j}) \; - (1-y) \log(1 - \gamma(x_i, x_j . p_{i,j}))$$

**Interaction edge target prediction.** Predicted interactions can then be used as input to the network for interaction target prediction. The features were obtained by concatenating node embeddings and dynamically calculated pairwise features. Finally, we trained the network with MSE loss.

$$L_\delta = MSE(\delta(x), y)$$

### 4.2. *evolver* module

The problem of having different targets for nodes with identical features can be solved by enforcing the permutation invariance in the loss function and letting the model make different predictions given identical features. The first part can be addressed by matching the model's predictions and target values. The second part can be solved by infusing randomness into the model (e.g., by randomly initializing hidden states for each node).

To compute the loss function value, we need to find a permutation with the lowest value of the loss function. Considering all possible permutations – even within one molecule – is completely unfeasible ($O(n!)$ time complexity, where $n$ is the number of nodes). To do that more efficiently, only lone pairs were selected. Then, within each group of lone pairs of a particular atom, a Hungarian algorithm was run to minimize the total loss function value for this group of nodes. This yields $O(gm^3)$ time complexity, where $g$ is the number of groups and $m$ is the number of nodes in the group). It is important to note, that $m$ is usually small: from 2 to 3 for the QM9 dataset, which makes the overall approach highly efficient. Finally, the corrected node order is used to compute final predictions (including predictions for pairwise tasks). This procedure is performed at each training step of the neural network.

All final predictions for the NBO model are based on the embeddings obtained from the graph neural network encoder. To infuse randomness, the most straightforward strategy is to initialize randomly sampled hidden states and then concatenate them with GNN embeddings to make final predictions. However, in such a setting, there is a chance that these embeddings will be similar enough to result in identical predictions. To solve that, we need to make the model aware of the existence of other nodes.

These ideas are reflected in the architecture of the evolver module (**Figure 2b**). First, concatenated node embeddings and hidden states are used to make node-level predictions. These predictions are then passed into the module along with corresponding hidden states. Then they are transformed into intermediate representations using multilayer fully connected networks (*A* and *B*, respectively). Representations of node embeddings and hidden states are then used to compute the dot product and corresponding weighting coefficients using a *softmax* function. These weights are then used to calculate new vectors, concatenated, and passed through another multiplayer fully connected decoding network (*C*). Finally, the decoded representation is added to the previous hidden states. The procedure is performed multiple times.

## Acknowledgments

The authors thank NSF ACCESS (project no. CHE220012), Google Cloud Platform, NVIDIA Academic Hardware Grant Program (project titled “New molecular graph representations in joint feature spaces”) for computational resources. G.G. and D.B. acknowledge the financial support by the National Science Foundation Center for Computer-Assisted Synthesis (Grant no. 2202693) and a supporting seed grant from X, the moonshot factory (an Alphabet company). G.G. thanks CMU and the departments of chemistry and chemical engineering for the startup support. G.G. thanks Prof. Frank Weinhold (UW Madison) for the development of NBO and the many discussions about the theory and software.

S.M.B. acknowledges financial support by the Laboratory Directed Research and Development Program of Lawrence Berkeley National Laboratory under U.S. Department of Energy Contract No. DE-AC02-05CH11231. Computational resources for the high-throughput virtual screening and datasets development were provided by the National Energy Research Scientific Computing Center (NERSC), a U.S. Department of Energy Office of Science User Facility under Contract No. DE-AC02-05CH11231, and by the Lawrencium computational cluster resource provided by the IT Division at the Lawrence Berkeley National Laboratory (Supported by the Director, Office of Science, Office of Basic Energy Sciences, of the U.S. Department of Energy under Contract No. DE-AC02-05CH11231).

We thank Prof. John Kitchin (CMU Chemical Engineering) and Prof. Olexandr Isayev (CMU Chemistry) for their constructive feedback.

Any opinions, findings, and conclusions or recommendations expressed in this material are those of the author(s) and do not necessarily reflect the views of the National Science

Foundation, the U.S. Department of Energy, Alphabet (and its subsidiaries), or any of the other funding sources.

## Code availability

The code is available at https://github.com/gomesgroup/simg.[64]

## Data availability

The data and model weights are available at https://huggingface.co/gomesgroup/simg.

## Author contributions

D.A.B. designed the computational pipeline and implemented ***SIMG**** prediction, active learning process, downstream task analysis and the first version of large molecule analysis. T.R. implemented the lone pair prediction model and performed analysis of large molecule predictions. B.S.-L. advised on the development of machine learning pipeline and software development. S.M.B. performed quantum chemistry calculations and advised on analysis of NBO data. G.G. designed the concept and performed preliminary studies. S.M.B. and G.G. supervised the project. D.A.B, T.R., and G.G. wrote this manuscript with input from all authors.

## References

1. Hoffmann, R. & Laszlo, P. Representation in Chemistry. *Angewandte Chemie International Edition in English* **30**, 1–16 (1991).

2. Cooke, H. A historical study of structures for communication of organic chemistry information prior to 1950. *Org Biomol Chem* **2**, 3179 (2004).

3. Springer, M. T. Improving Students' Understanding of Molecular Structure through Broad-Based Use of Computer Models in the Undergraduate Organic Chemistry Lecture. *J Chem Educ* **91**, 1162–1168 (2014).

4. Gómez-Bombarelli, R. *et al.* Design of efficient molecular organic light-emitting diodes by a high-throughput virtual screening and experimental approach. *Nat Mater* **15**, 1120–1127 (2016).

5. Zhavoronkov, A. *et al.* Deep learning enables rapid identification of potent DDR1 kinase inhibitors. *Nat Biotechnol* **37**, 1038–1040 (2019).

6. Dara, S., Dhamercherla, S., Jadav, S. S., Babu, C. M. & Ahsan, M. J. Machine Learning in Drug Discovery: A Review. *Artif Intell Rev* **55**, 1947–1999 (2022).

7. Gallegos, L. C., Luchini, G., St. John, P. C., Kim, S. & Paton, R. S. Importance of Engineered and Learned Molecular Representations in Predicting Organic

Reactivity, Selectivity, and Chemical Properties. *Acc Chem Res* **54**, 827–836 (2021).

8. Sandfort, F., Strieth-Kalthoff, F., Kühnemund, M., Beecks, C. & Glorius, F. A Structure-Based Platform for Predicting Chemical Reactivity. *Chem* **6**, 1379–1390 (2020).

9. Ross, J. *et al.* Large-scale chemical language representations capture molecular structure and properties. *Nat Mach Intell* **4**, 1256–1264 (2022).

10. Yang, Z., Chakraborty, M. & White, A. D. Predicting chemical shifts with graph neural networks. *Chem Sci* **12**, 10802–10809 (2021).

11. Zhou, J. *et al.* Graph neural networks: A review of methods and applications. *AI Open* **1**, 57–81 (2020).

12. Fang, X. *et al.* Geometry-enhanced molecular representation learning for property prediction. *Nat Mach Intell* **4**, 127–134 (2022).

13. Batzner, S. *et al.* E(3)-equivariant graph neural networks for data-efficient and accurate interatomic potentials. *Nat Commun* **13**, 2453 (2022).

14. Qi, Y., Gong, W. & Yan, Q. Bridging deep learning force fields and electronic structures with a physics-informed approach. (2024).

15. Fabrizio, A., Briling, K. R. & Corminboeuf, C. SPA $^{H}$ M: the spectrum of approximated Hamiltonian matrices representations. *Digital Discovery* **1**, 286–294 (2022).

16. Elton, D. C., Boukouvalas, Z., Butrico, M. S., Fuge, M. D. & Chung, P. W. Applying machine learning techniques to predict the properties of energetic materials. *Sci Rep* **8**, 9059 (2018).

17. Rupp, M., Tkatchenko, A., Müller, K.-R. & von Lilienfeld, O. A. Fast and Accurate Modeling of Molecular Atomization Energies with Machine Learning. *Phys Rev Lett* **108**, 058301 (2012).

18. Pozdnyakov, S. N. & Ceriotti, M. Smooth, exact rotational symmetrization for deep learning on point clouds. *ArXiv 2305.19302* (2023).

19. Abramson, J. *et al.* Accurate structure prediction of biomolecular interactions with AlphaFold 3. *Nature* **630**, 493–500 (2024).

20. Černý, J. & Hobza, P. Non-covalent interactions in biomacromolecules. *Physical Chemistry Chemical Physics* **9**, 5291 (2007).

21. Anighoro, A. Underappreciated Chemical Interactions in Protein–Ligand Complexes. in *Quantum Mechanics in Drug Discovery* 75–86 (2020). doi:10.1007/978-1-0716-0282-9_5.

22. Wheeler, S. E., Seguin, T. J., Guan, Y. & Doney, A. C. Noncovalent Interactions in Organocatalysis and the Prospect of Computational Catalyst Design. *Acc Chem Res* **49**, 1061–1069 (2016).

23. WEINHOLD, F. & LANDIS, C. R. NATURAL BOND ORBITALS AND EXTENSIONS OF LOCALIZED BONDING CONCEPTS. *Chem. Educ. Res. Pract.* **2**, 91–104 (2001).

24. Llenga, S. & Gryn'ova, G. Matrix of orthogonalized atomic orbital coefficients representation for radicals and ions. *J Chem Phys* **158**, (2023).

25. Bartók, A. P., Kondor, R. & Csányi, G. On representing chemical environments. *Phys Rev B* **87**, 184115 (2013).

26. Prokhorenkova, L., Gusev, G., Vorobev, A., Dorogush, A. V. & Gulin, A. Catboost: Unbiased boosting with categorical features. *Adv Neural Inf Process Syst* **2018-Decem**, 6638–6648 (2018).

27. NVIDIA. MegaMolBART. *GitHub* https://github.com/NVIDIA/MegaMolBART.

28. Heid, E. *et al.* Chemprop: A Machine Learning Package for Chemical Property Prediction. *J Chem Inf Model* **64**, 9–17 (2024).

29. Alabugin, I. V. *Stereoelectronic Effects: A Bridge Between Structure and Reactivity*. (John Wiley & Sons, 2016).

30. Echenique, P. & Alonso, J. L. A mathematical and computational review of Hartree–Fock SCF methods in quantum chemistry. *Mol Phys* **105**, 3057–3098 (2007).

31. Burke, K. & Wagner, L. O. DFT in a nutshell. *Int J Quantum Chem* **113**, 96–101 (2013).

32. Goerigk, L. & Grimme, S. Double-hybrid density functionals. *Wiley Interdiscip Rev Comput Mol Sci* **4**, 576–600 (2014).

33. Kneiding, H. *et al.* Deep learning metal complex properties with natural quantum graphs. *Digital Discovery* **2**, 618–633 (2023).

34. Johnson, E. R. *et al.* Revealing Noncovalent Interactions. *J Am Chem Soc* **132**, 6498–6506 (2010).

35. Axelrod, S. & Gómez-Bombarelli, R. GEOM, energy-annotated molecular conformations for property prediction and molecular generation. *Sci Data* **9**, 185 (2022).

36. Malinin, A., Prokhorenkova, L. & Ustimenko, A. Uncertainty in Gradient Boosting via Ensembles. *ArXiv 2006.10562* 1–17 (2020).

37. Chua, K., Calandra, R., McAllister, R. & Levine, S. Deep Reinforcement Learning in a Handful of Trials using Probabilistic Dynamics Models. (2018).

38. Goan, E. & Fookes, C. Bayesian Neural Networks: An Introduction and Survey. (2020) doi:10.1007/978-3-030-42553-1_3.

39. Beluch, W. H., Genewein, T., Nurnberger, A. & Kohler, J. M. The Power of Ensembles for Active Learning in Image Classification. in *2018 IEEE/CVF Conference on Computer Vision and Pattern Recognition* 9368–9377 (IEEE, 2018). doi:10.1109/CVPR.2018.00976.

40. León, I., Alonso, E. R., Cabezas, C., Mata, S. & Alonso, J. L. Unveiling the n→π* interactions in dipeptides. *Commun Chem* **2**, 3 (2019).

41. Newberry, R. W., Bartlett, G. J., VanVeller, B., Woolfson, D. N. & Raines, R. T. Signatures of *n→π** interactions in proteins. *Protein Science* **23**, 284–288 (2014).

42. Hodges, J. A. & Raines, R. T. Energetics of an *n → π* * Interaction that Impacts Protein Structure. *Org Lett* **8**, 4695–4697 (2006).

43. Zhou, Y., Morais-Cabral, J. H., Kaufman, A. & MacKinnon, R. Chemistry of ion coordination and hydration revealed by a K+ channel–Fab complex at 2.0 Å resolution. *Nature* **414**, 43–48 (2001).

44. Bartlett, G. J., Choudhary, A., Raines, R. T. & Woolfson, D. N. n→π* interactions in proteins. *Nat Chem Biol* **6**, 615–620 (2010).

45. dos Passos Gomes, G. & Alabugin, I. V. Drawing Catalytic Power from Charge Separation: Stereoelectronic and Zwitterionic Assistance in the Au(I)-Catalyzed Bergman Cyclization. *J Am Chem Soc* **139**, 3406–3416 (2017).

46. Gomes, G. dos P., Vil', V., Terent'ev, A. & Alabugin, I. V. Stereoelectronic source of the anomalous stability of bis-peroxides. *Chem Sci* **6**, 6783–6791 (2015).

47. Grabowski, S. J. Tetrel bond–σ-hole bond as a preliminary stage of the $S_N2$ reaction. *Phys. Chem. Chem. Phys.* **16**, 1824–1834 (2014).

48. Sarazin, Y., Liu, B., Roisnel, T., Maron, L. & Carpentier, J.-F. Discrete, Solvent-Free Alkaline-Earth Metal Cations: Metal⋯Fluorine Interactions and ROP Catalytic Activity. *J Am Chem Soc* **133**, 9069–9087 (2011).

49. Ramakrishnan, R., Dral, P. O., Rupp, M. & von Lilienfeld, O. A. Quantum chemistry structures and properties of 134 kilo molecules. *Sci Data* **1**, 140022 (2014).

50. Mardirossian, N. & Head-Gordon, M. $\omega$ B97M-V: A combinatorially optimized, range-separated hybrid, meta-GGA density functional with VV10 nonlocal correlation. *J Chem Phys* **144**, 214110 (2016).

51. Shao, Y. *et al.* Advances in molecular quantum chemistry contained in the Q-Chem 4 program package. *Mol Phys* **113**, 184–215 (2015).

52. Glendening, E. D., Landis, C. R. & Weinhold, F. *NBO 7.0* : New vistas in localized and delocalized chemical bonding theory. *J Comput Chem* jcc.25873 (2019) doi:10.1002/jcc.25873.

53. Ong, S. P. *et al.* Python Materials Genomics (pymatgen): A robust, open-source python library for materials analysis. *Comput Mater Sci* **68**, 314–319 (2013).

54. Blau, S., Spotte-Smith, E. W. C., Wood, B., Dwaraknath, S. & Persson, K. Accurate, Automated Density Functional Theory for Complex Molecules Using On-the-fly Error Correction. *ChemRxiv* (2020) doi:10.26434/chemrxiv.13076030.v1.

55. Mathew, K. *et al.* Atomate: A high-level interface to generate, execute, and analyze computational materials science workflows. *Comput Mater Sci* **139**, 140–152 (2017).

56. Paszke, A. *et al.* PyTorch: An Imperative Style, High-Performance Deep Learning Library. in *Advances in Neural Information Processing Systems 32* (eds. Wallach, H. et al.) 8024–8035 (Curran Associates, Inc., 2019).

57. Falcon, W. A., et al. PyTorch Lightning. *GitHub. Note: https://github.com/PyTorchLightning/pytorch-lightning* (2019).

58. Fey, M. & Lenssen, J. E. Fast Graph Representation Learning with PyTorch Geometric. *ArXiv 1903.02428* (2019).

59. Corso, G., Cavalleri, L., Beaini, D., Liò, P. & Veličković, P. Principal Neighbourhood Aggregation for Graph Nets. *ArXiv 2004.05718* (2020).

60. Li, G., Müller, M., Thabet, A. & Ghanem, B. DeepGCNs: Can GCNs Go as Deep as CNNs? *ArXiv 1904.03751* (2019).

61. Godwin, J. *et al.* Simple GNN Regularisation for 3D Molecular Property Prediction & Beyond. *ArXiv 2106.07971* (2021).

62. Veličković, P. *et al.* Graph Attention Networks. *ArXiv 1710.10903* (2017).

63. Cai, C. & Wang, Y. A Note on Over-Smoothing for Graph Neural Networks. *ArXiv 2006.13318* (2020).

64. Advancing Molecular Machine Learned Representations with Stereoelectronics-Infused Molecular Graphs. Preprint at https://doi.org/10.5281/zenodo.14393496.

*Supporting information for the article*

# Advancing Molecular Machine Learning Representations with Stereoelectronics-Infused Molecular Graphs

*by*

Daniil A. Boiko,[1] Thiago Reschützegger,[2] Benjamin Sanchez-Lengeling,[3,4,5] Samuel M. Blau,*[6] Gabe Gomes*[1,7,8,9]

1. Department of Chemical Engineering, Carnegie Mellon University, Pittsburgh, PA 15213, USA
2. Department of Chemical Engineering, Federal University of Santa Maria, Santa Maria, RS, Brazil
3. Google DeepMind, Cambridge, MA, USA (*previous affiliation, where most of this work was done*)
4. Department of Chemical Engineering and Applied Chemistry, University of Toronto, Toronto, ON M5S 3E5, Canada (*current affiliation*)
5. Vector Institute for Artificial Intelligence, Toronto, ON, Canada (*current affiliation*)
6. Energy Technologies Area, Lawrence Berkeley National Laboratory, Berkeley, CA 94720, USA
7. Department of Chemistry, Carnegie Mellon University, Pittsburgh, PA 15213, USA
8. Machine Learning Department, Carnegie Mellon University, Pittsburgh, PA 15213, USA
9. Wilton E. Scott Institute for Energy Innovation, Carnegie Mellon University, Pittsburgh, PA 15213, USA

** corresponding authors: smblau@lbl.gov, gabegomes@cmu.edu*

# SIMG prediction model

## General notes

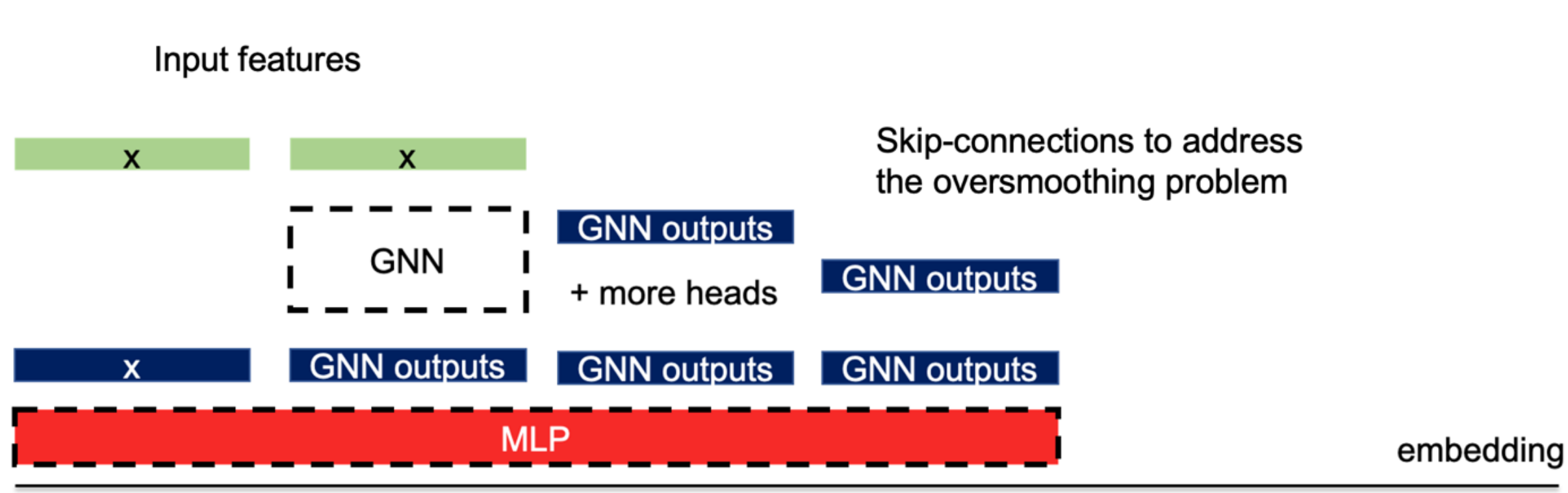

**Figure S1.** General overview of the architecture. Skip-connections were added to tackle over smoothing problem.

## *evolver* module

### Node reordering process

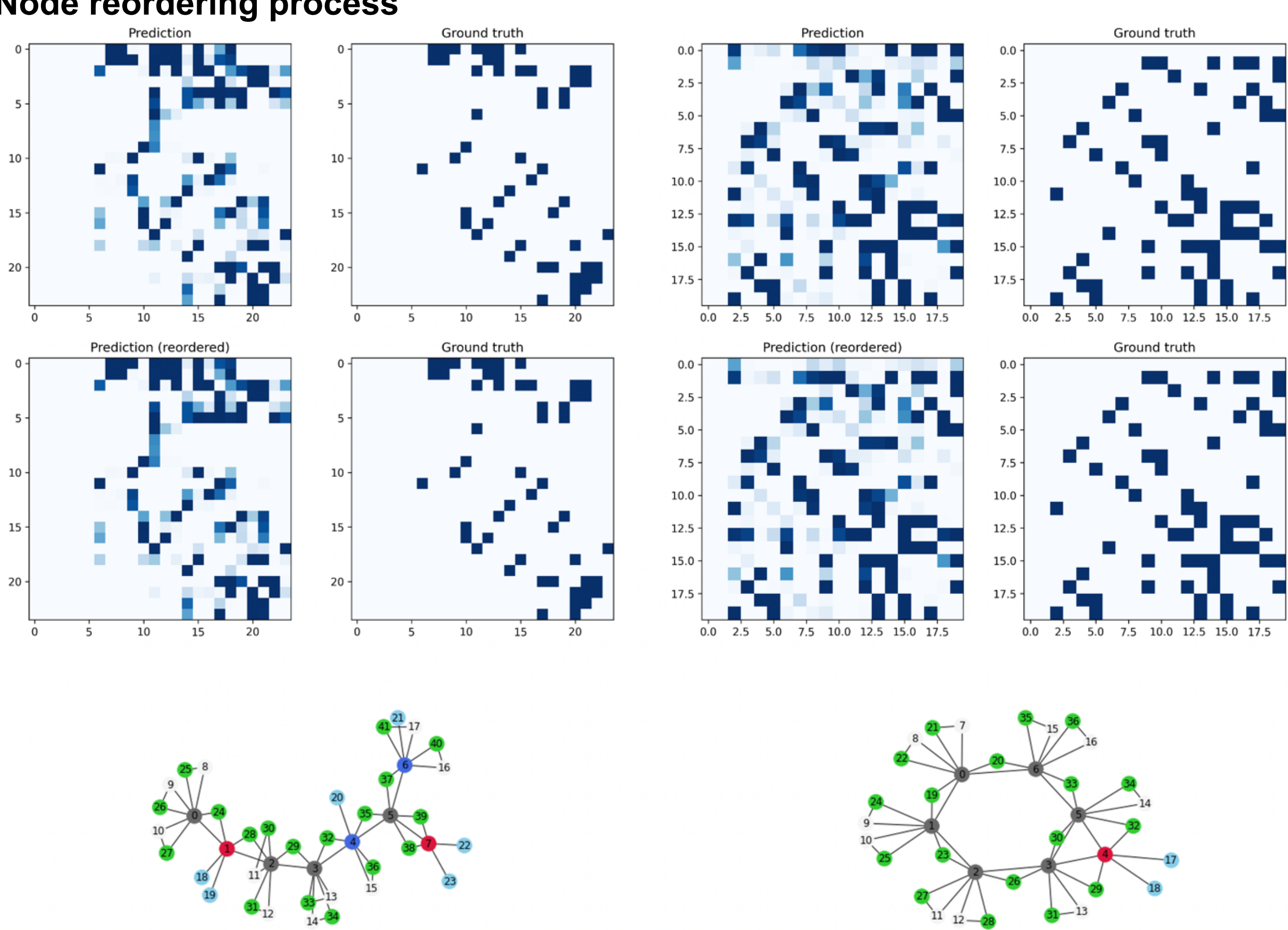

**Figure S2.** Overview of the node reordering procedure results for one of checkpoints for ***evolver***-based model. Graphs below represent input extended molecular graphs.

### Permutation-invariant loss function

The `TotalLoss` class defines a custom loss function for a neural network model that processes graph structures, focusing particularly on preserving permutation invariance among node groupings. The class is structured to compute multiple types of losses: node loss via `NodeLoss`, link prediction loss via binary cross-entropy (`BCEWithLogitsLoss`), atom-to-bond loss via mean squared error (`MSELoss`), and interaction loss. One significant feature of `TotalLoss` is its optional node matching step, controlled by the `perform_matching` flag. If enabled, this step reorders nodes according to a matching obtained from the `get_matching` function, which ensures that the node predictions (`node_preds`) are aligned with the targets (`node_targets`) based on their similarity. This is critical for learning graph representations where the ordering of nodes within groups can vary but should not affect the model's outcome — hence ensuring permutation invariance. The remaining calculations adjust predictions and target tensors for links and interactions based on these matchings, filtering out affected components to properly align paired elements before the final computation of the combined loss using the individual losses. This approach ensures that the loss computation respects the inherent permutation invariance required in many graph-based applications, making the model robust to node reorderings within identified groups of lone pairs.

## Examples of application

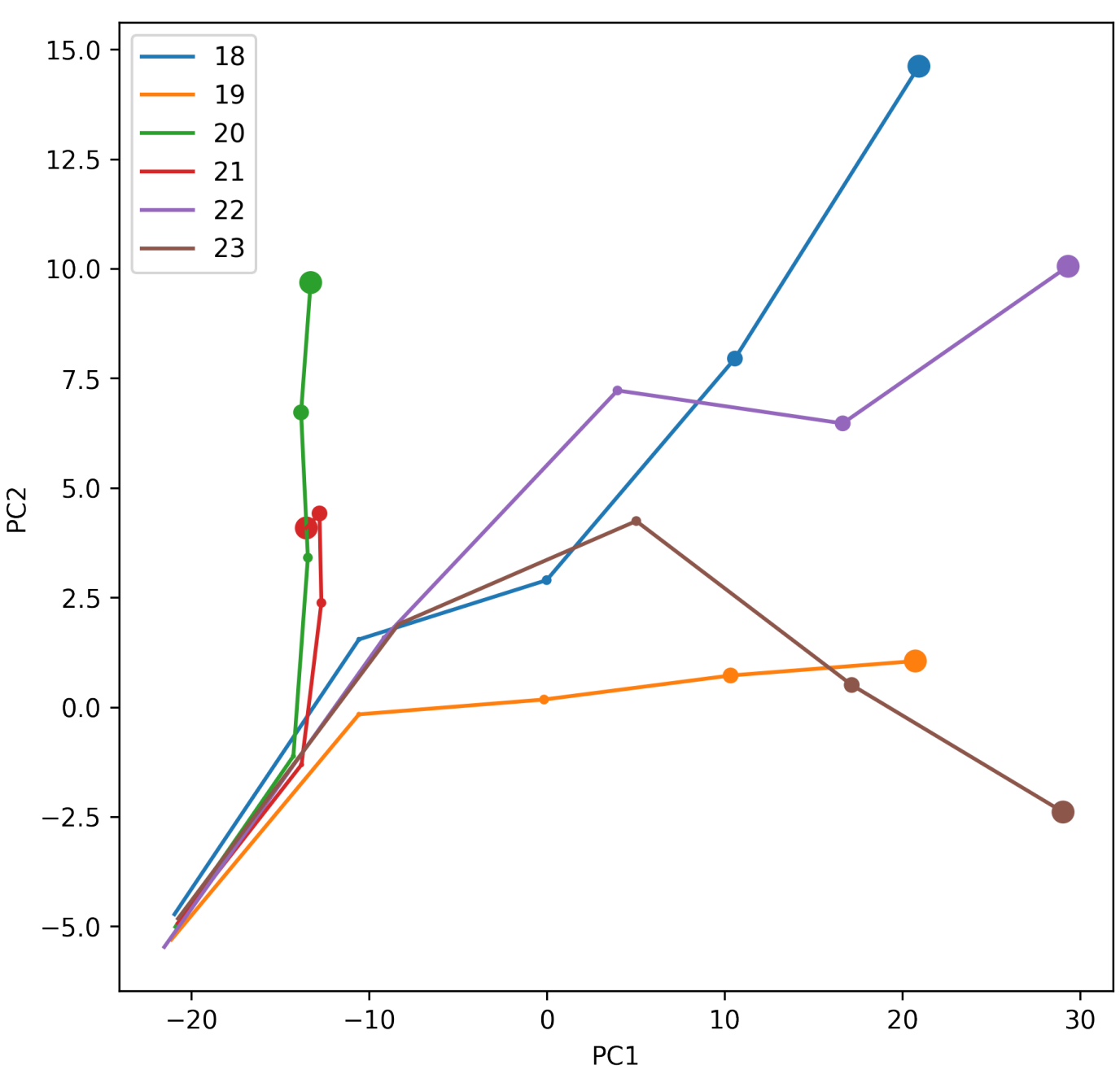

**Figure S3.** Example changes of lone positions in PCA space over multiple iterations. Point size indicates step.

## Results in comparison to other methods

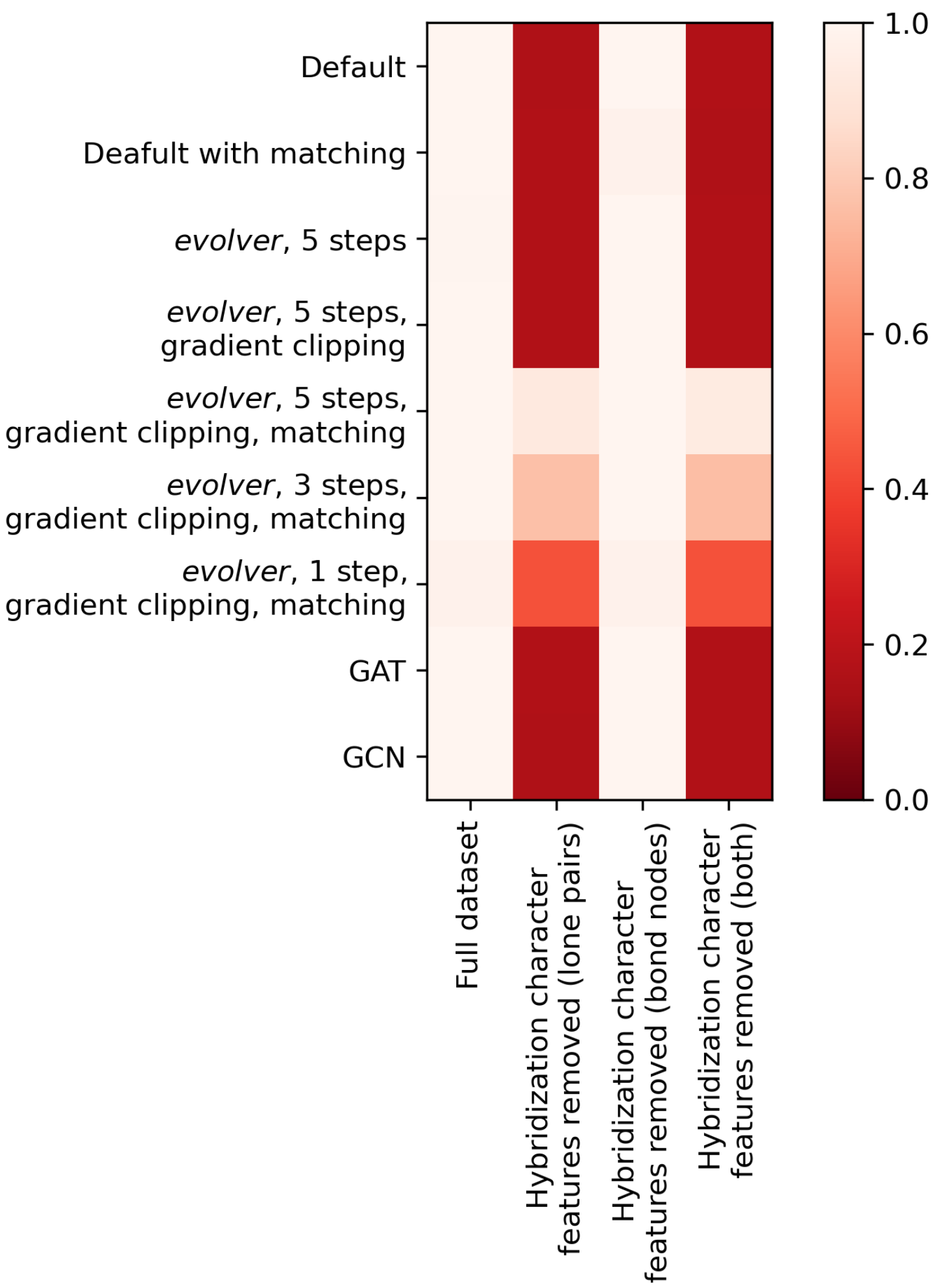

**Figure S4.** Results for lone pair prediction model (*s*-character)

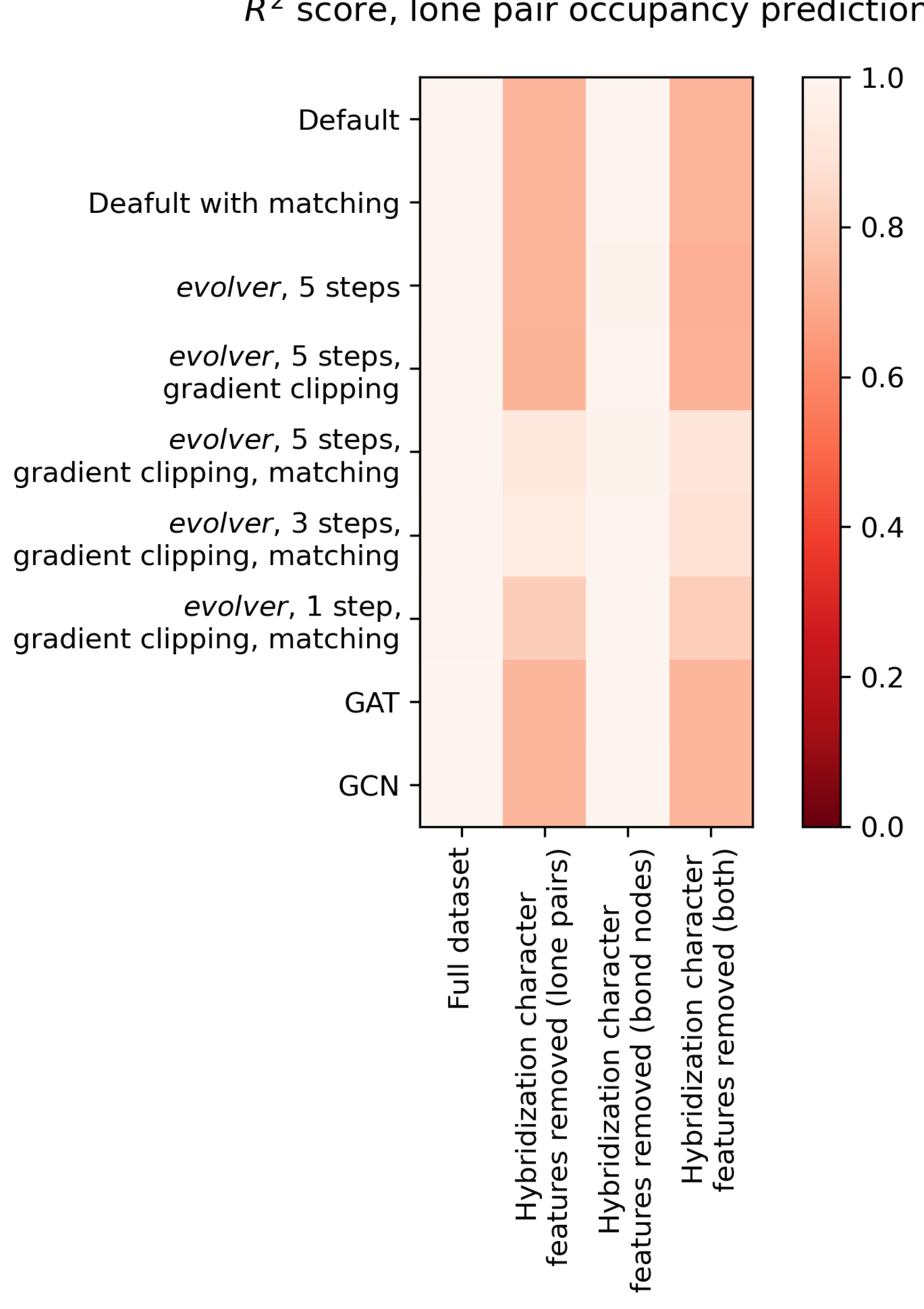

**Figure S5.** Results for lone pair prediction model (occupancy)

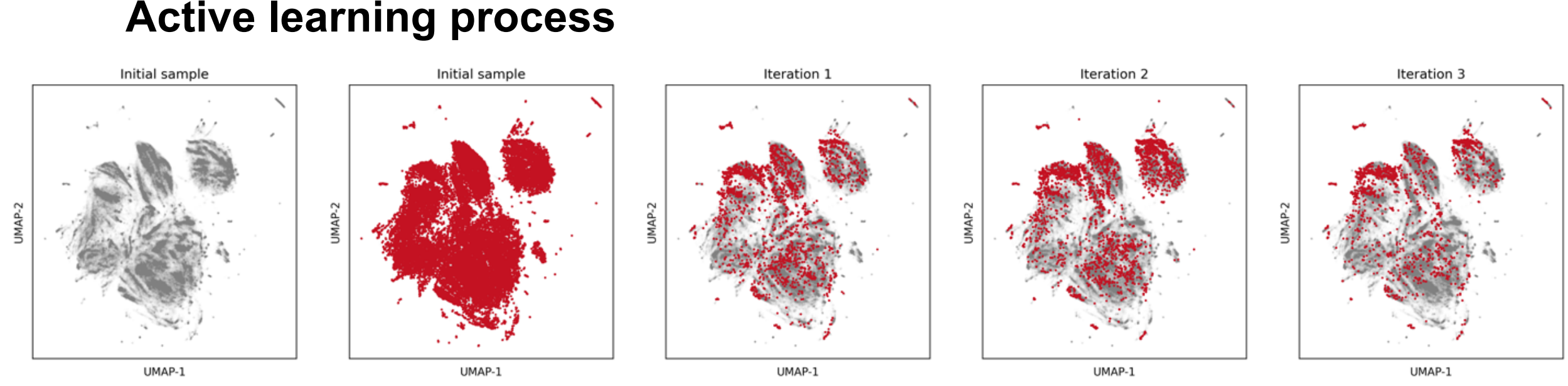

**Figure S6.** UMAP plot of chemical space navigation during active learning procedure.

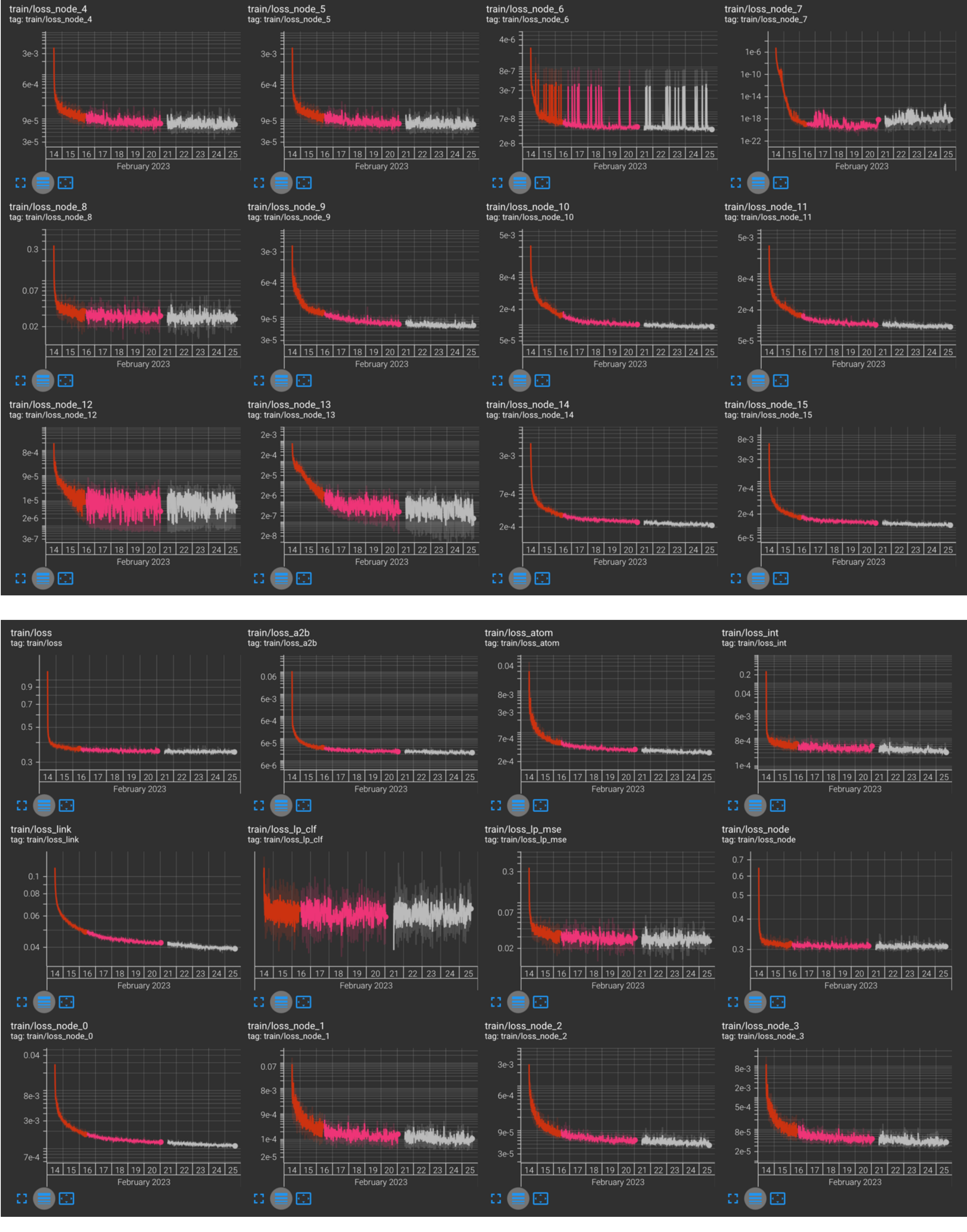

**Figure S7.** Training loss curves for the final model training run.

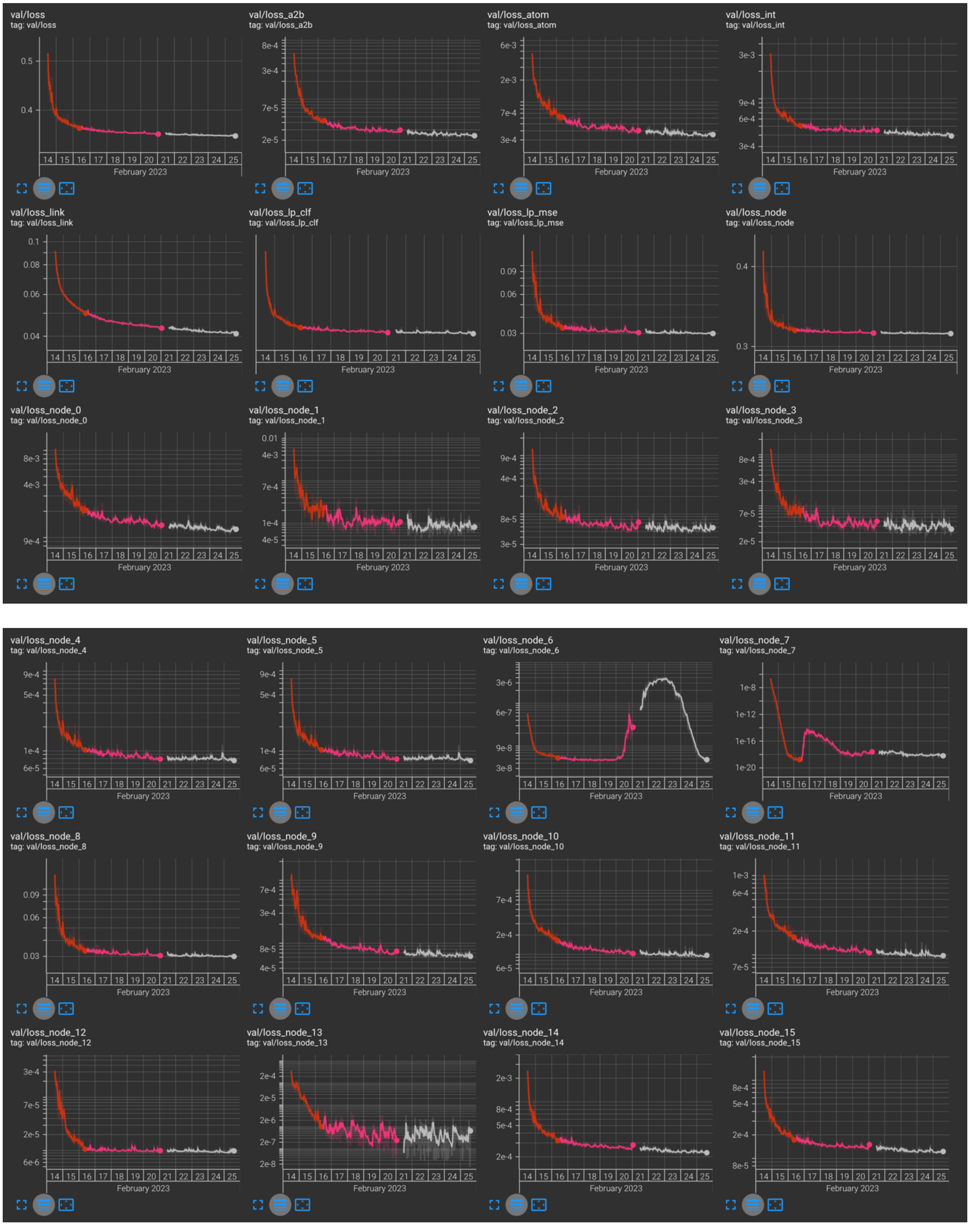

**Figure S8.** Validation loss curves for the final model training run.

**Baseline node embedding space**

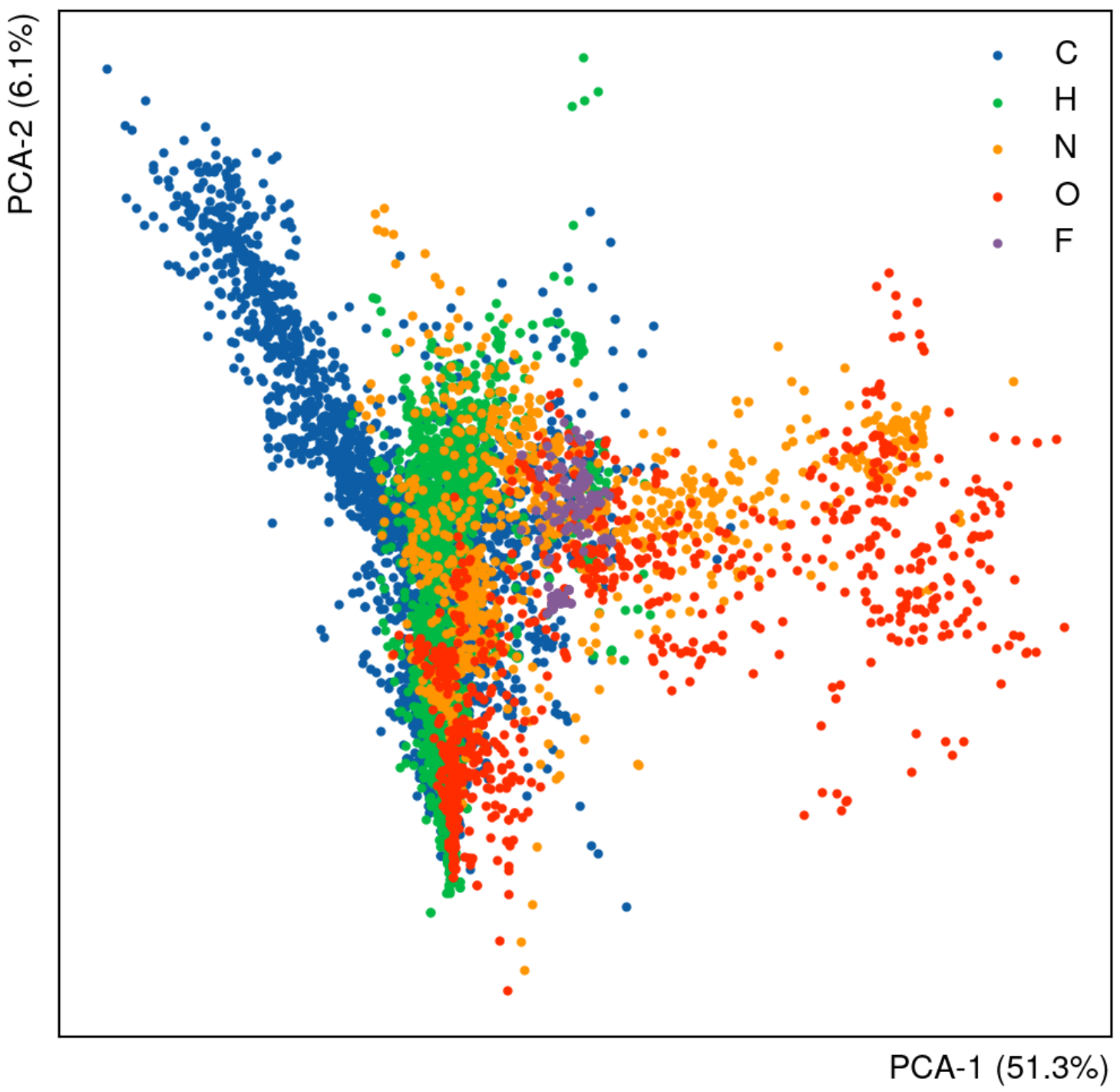

**Figure S9.** Node embedding space for GNN model, trained to predict QM9 targets.

# Protein SIMG* prediction performance

Interaction matrices depict the connections between orbital nodes in SIMGs, organized by their respective indices *i* and *j*. An entry of 1 at position (*i*, *j*) indicates an interaction where electron density is transferred from donor *j* to acceptor *i*. The indices are arranged such that lone-pair (LP) nodes precede bond nodes. Red areas in the matrix signify non-existent interactions between lone pairs of electrons. Purple areas denote interactions between bond orbitals and lone pairs, while blue areas illustrate interactions among bond orbitals.

Figure S10 displays the interaction matrices for both the predicted and ground-truth graphs, as well as their absolute differences, for the out-of-distribution protein with PDB code 1k4d. The model demonstrates high accuracy in predicting a large number of interactions and correctly identifies the absence of interactions between lone pairs or electron density transfer from bond orbitals to electron-rich lone-pair orbitals. The threshold used for these predictions was set at 0.95.

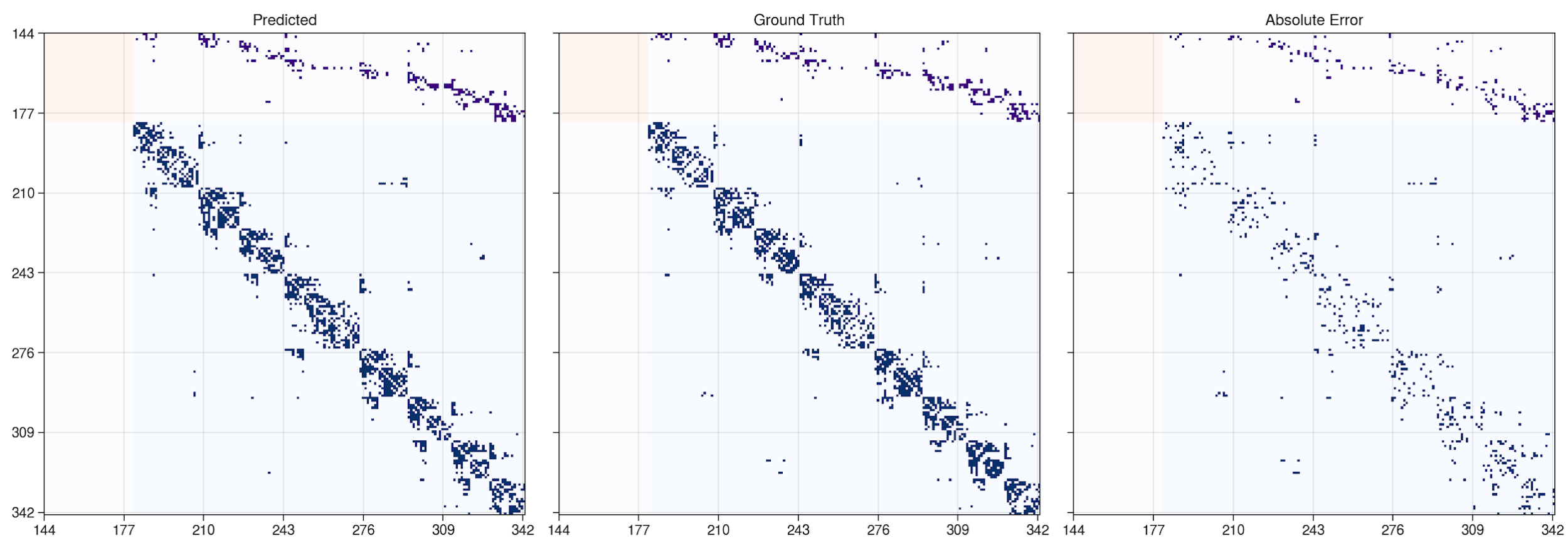

**Figure S10.** Predicted, ground-truth interactions matrices for protein PDB code 1k4d, followed by the absolute error.

Moreover, the model accurately predicts properties for this molecule. Proving that it is able to generalize to arbitrary long, out-of-distribution, molecules. Figure 11a, b, and c show predictions for node, lone-pair, and bond orbitals, respectively.

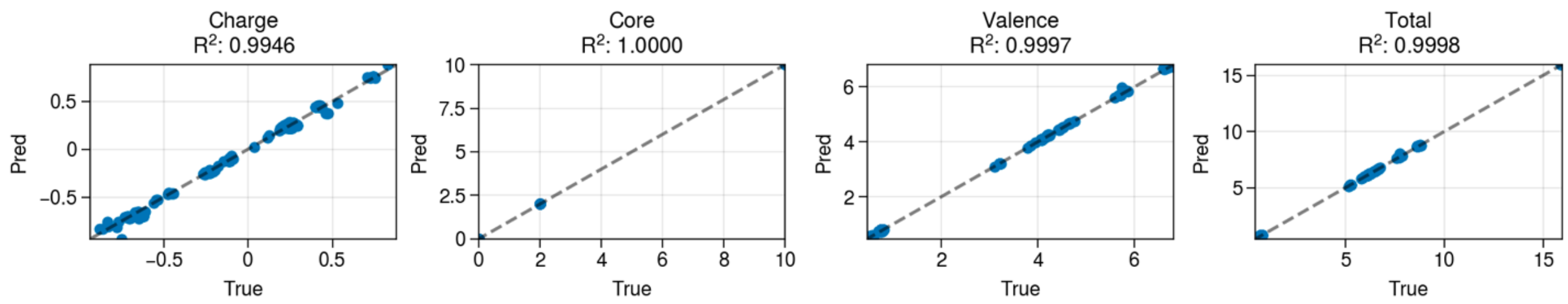

**Figure S11a.** Predicted atom node targets.

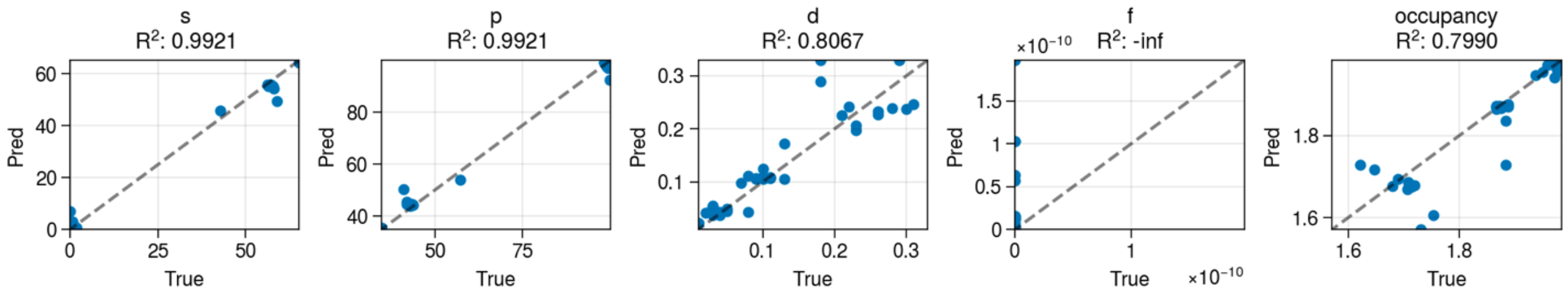

**Figure S11b.** Predicted lone-pair node targets.

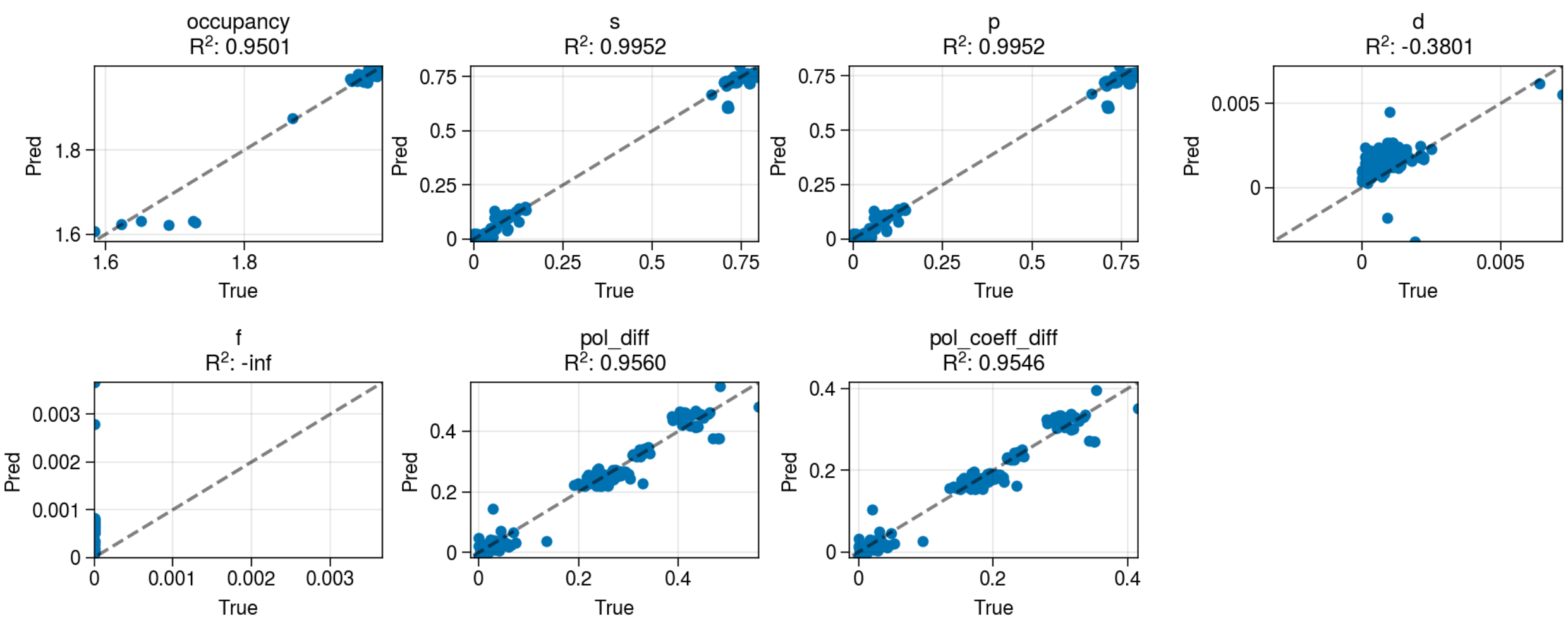

**Figure S11c.** Predicted bond orbital node targets.

# Downstream property prediction models

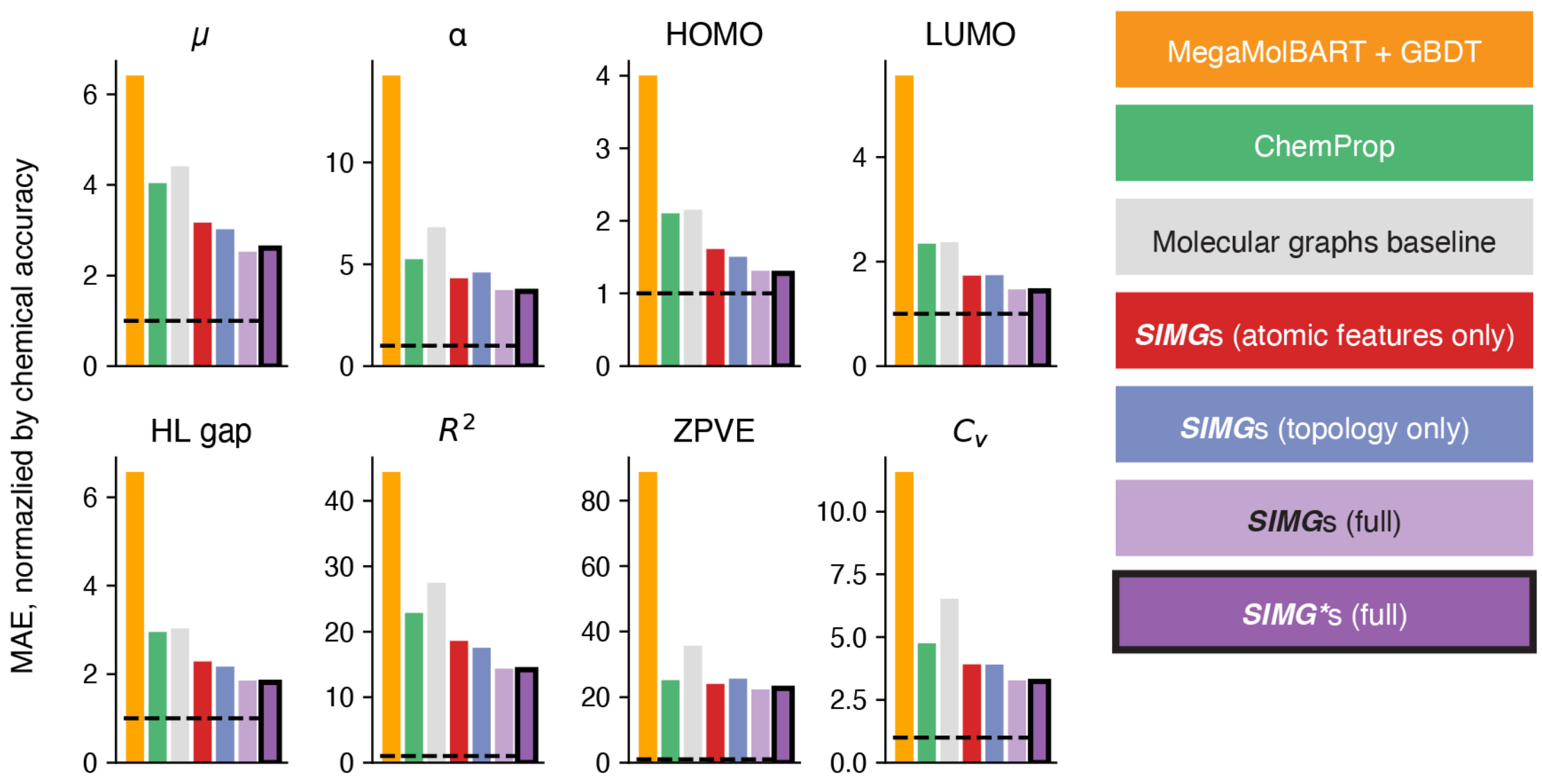

**Figure S12.** Comparison with MegaMolBART.